\pdfoutput=1

\documentclass[11pt]{article}

\usepackage[preprint]{acl}

\usepackage{times}
\usepackage{latexsym}
\usepackage{tcolorbox}
\usepackage{parskip} 
\usepackage[T1]{fontenc}

\usepackage[utf8]{inputenc}
\usepackage{flushend}
\usepackage{microtype}

\usepackage{inconsolata}

\usepackage{graphicx}

\usepackage{float} 
\usepackage{amsmath}
\usepackage{pifont}
\usepackage{array}
\usepackage{booktabs}
\usepackage{enumitem}
\usepackage{amssymb}
\usepackage{multicol}
\newcommand{\vpara}[1]{\vspace{0.05in}\noindent \textbf{#1 }}
\newcommand{\model}{FacLens}
\newcommand{\smodel}{FacLens\space}
\newtheorem{problem}{\textbf{Problem}}
\newtheorem{definition}{\textbf{Definition}}

\newtheorem{remark}{Remark}

\usepackage{threeparttable}
\usepackage{makecell}
\usepackage{graphicx}
\usepackage{multirow}
\usepackage{graphicx}
\usepackage{subfigure}
\usepackage{wrapfig}
\usepackage{graphicx}
\usepackage{caption}

\usepackage{hyperref}
\hypersetup{
    colorlinks = true,
    citecolor = {blue}, 
    urlcolor = {blue},
    linkcolor = {blue}  
}
\usepackage{tcolorbox}

\title{\model: Transferable Probe for Foreseeing Non-Factuality in Fact-Seeking Question Answering of Large Language Models}

\author{
Yanling Wang$^{1}$\thanks{Work done while at Zhongguancun Laboratory.},
Haoyang Li$^{3}$,
Hao Zou$^{2}$,
Jing Zhang$^{3}$\thanks{Corresponding authors.},
Xinlei He$^{4}$,
Qi Li$^{2,5}$\footnotemark[2],
Ke Xu$^{2,5}$\\
$^{1}$Zhipu AI \quad
$^{2}$Zhongguancun Laboratory \quad
$^{3}$Renmin University of China\\
$^{4}$Hong Kong University of Science and Technology (Guangzhou) \quad
$^{5}$Tsinghua University\\
\texttt{yanlingwang777@gmail.com} \quad
\texttt{zhang-jing@ruc.edu.cn} \quad
\texttt{qli01@tsinghua.edu.cn}
}

\begin{document}
\maketitle
\begin{abstract}
Despite advancements in large language models (LLMs), non-factual responses still persist in fact-seeking question answering. Unlike extensive studies on post-hoc detection of these responses, this work studies non-factuality prediction (NFP), predicting whether an LLM will generate a non-factual response prior to the response generation.
Previous NFP methods have shown LLMs' awareness of their knowledge, but they face challenges in terms of efficiency and transferability.
In this work, we propose a lightweight model named \textbf{Fac}tuality \textbf{Lens} (\model), which effectively probes hidden representations of fact-seeking questions for the NFP task. Moreover, we discover that hidden question representations sourced from different LLMs exhibit similar NFP patterns, enabling the transferability of \smodel across different LLMs to reduce development costs.
Extensive experiments highlight \model's superiority in both effectiveness and efficiency.~\footnote{Code: \url{https://github.com/wyl7/FacLens}.}
\end{abstract}

\section{Introduction}
\begin{figure*}
	\centering
		\includegraphics[width=5.3in]{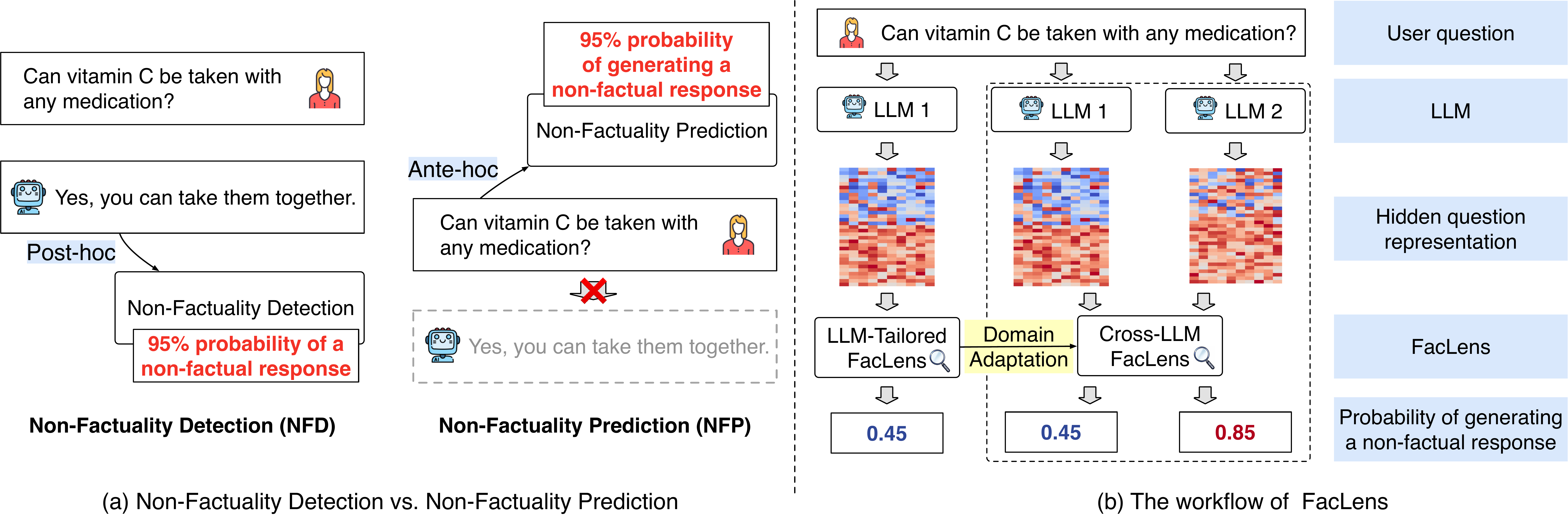}
      \caption{Illustrations of the objective and workflow of \smodel for efficient and transferable ante-hoc NFP.}
  \label{fig:into_target_of_this_paper}
\end{figure*}

Large language models (LLMs) have shown impressive abilities in understanding and generating coherent text~\citep{GPT4TechnicalReport, llama3, mistral}, yet they may provide non-factual responses in fact-seeking question answering (fact-seeking QA)~\citep{zhang2023siren, riskTaxonomy2024}.
Extensive studies have been devoted to detecting the non-factual responses, a task we name {non-factuality detection} (NFD)~\citep{SelfCheckGPT,SAPLMA,INSIDE,chen2023complex, min2023factscore}.
However, these post-hoc methods require response generation, which incurs significant computational overhead.
Therefore, this paper studies {non-factuality prediction} (NFP), which predicts the likelihood of an LLM generating a non-factual response to a fact-seeking question before the response generation.
Figure~\ref{fig:into_target_of_this_paper} (a) illustrates the difference between NFD and NFP.

To solve the NFP problem, researchers have proposed predicting by analyzing specific tokens in a question~\citep{whenToUseExternalKnowledge, SATProbe}, making these methods applicable to specific types of questions. For more general questions, approaches based on prompting or fine-tuning the LLM for NFP have been proposed~\cite{SelfFAMILIARITY, LLMKnowWhatTheyKnow}. However, two limitations persist: (1) current NFP models can be improved in terms of effectiveness and efficiency, and (2) they are designed for an individual LLM, lacking the transferability for rapid adaptation to new LLMs.

Inspired by studies that monitor and manipulate hidden representations to improve LLMs' performance~\citep{RepE, TruthX, ProbeEntityforMitigateHallu}, we hypothesize that knowledge awareness has been embedded in the hidden representations of fact-seeking questions. To investigate this, we propose a lightweight model, \textbf{Fac}tuality \textbf{Lens} (\model), and demonstrate its ability to probe the hidden question representations for NFP.
Figure~\ref{fig:into_target_of_this_paper} (b) shows the workflow of \model. 
As hidden question representations can be efficiently obtained, and the model structure is lightweight, \smodel achieves high \textbf{efficiency in both training and prediction} (see Table~\ref{tab:efficacy}). This makes it highly suitable for real-world LLM applications, helping to reduce non-factual responses to end-users while maintaining a minimal budget and latency.
To construct the training data of \model, we prompt the target LLM to produce responses to questions from high-quality fact-seeking datasets. We then compare the LLM-generated answers with the golden answers, assigning binary factual/non-factual labels to the responses. Nevertheless, extending \smodel to support multiple LLMs becomes resource-intensive and time-consuming, because each LLM must conduct response generation for the training data construction. Fortunately, we discover the transferability of \smodel across different LLMs, allowing us to assign the binary labels for just one LLM and apply unsupervised domain adaptation (DA) to rapidly apply \smodel to other LLMs without collecting new labels, thereby improving the \textbf{efficiency in development}.

Investigating the transferability of \smodel is inspired by the research on human cognition~\citep{miller2012individual}, which shows that individuals with similar cognitive styles and encoding strategies exhibit similar brain activity when performing the same task. 
Given current LLMs generally follow the Transformer~\citep{transformer} architecture and share overlapping training corpora~\cite{gao2020pile,commoncrawl,kocetkov2022stack}, we hypothesize that different LLMs have similar cognitive patterns in terms of knowledge awareness (i.e., similar NFP patterns).
To validate this hypothesis, we define a collection of hidden question representations sourced from a certain LLM as a data domain.
Our experiments demonstrate that a \smodel trained on data from multiple domains performs similarly to the one trained on a single domain, suggesting that different individual domains do not have a significant concept shift~\citep{conceptshift}.\footnote{Not having a significant concept shift implies highly consistent conditional distributions $P(\mathbf{y}|\mathbf{X})$ between domains.}
Building on this insight, we can quickly apply a trained \smodel to a new LLM through unsupervised domain adaptation (DA)~\citep{RKHSforDistributions, unsupervisedDomainAdaptation}. 
Additionally, we propose a question-aligned strategy to enhance the mini-batch-based DA.

This paper makes the following contributions: 
\begin{itemize}[leftmargin=2em]
    \item \textbf{Findings.} We 
    verify that hidden question representations in an LLM contain valuable information for NFP in fact-seeking QA, i.e., LLMs' activation during question-embedding mostly reveals whether they know the facts.
    Moreover, we show that different LLMs exhibit similar NFP patterns to enable a cross-LLM NFP model.
    \item  \textbf{Method.}
    We propose a lightweight and transferable NFP model named \model, enabling efficient development and application. To our knowledge, this is a pioneer work to train a transferable NFP model for multiple LLMs.
    \item \textbf{Experiments.} We conduct extensive experiments, involving four open-source LLMs and three widely used fact-seeking QA datasets. The results show that \smodel outperforms baselines in terms of AUC metric and runtime.
\end{itemize}

\section{Related Work}\label{sec:related}
\vpara{Hidden Representations in LLMs.}
Hidden representations in LLMs have been shown to encode valuable information that can be leveraged for various tasks~\cite{RepE}.
In terms of LLMs' factuality, studies like SAPLMA~\cite{SAPLMA} and MIND~\cite{MIND} leverage hidden representations of LLM-generated responses for post-hoc NFD.
TruthX~\cite{TruthX} edits hidden representations of LLM-generated responses via an edit vector to enhance the LLM's truthfulness.
Activation Decoding~\cite{ProbeEntityforMitigateHallu} reveals that an LLM's responses are closely tied to the representations of input entities but do not validate their effectiveness in NFP.

\vpara{Non-Factuality Prediction in LLMs.}
We categorize the studies into token-based and non-token-based methods.
The entity popularity-based method~\cite{whenToUseExternalKnowledge} focuses on input entity tokens, assuming that LLMs are more familiar with questions about popular entities and estimating entity popularity based on Wikipedia page views. However, not every question contains entities that exactly match a Wikipedia entry. SAT Probe~\cite{SATProbe} predicts based on the LLM's attention to specific constraint tokens. 
The authors restrict the types and formats of questions to facilitate the identification of the constraint tokens. However, extracting constraint tokens from free-form questions is non-trivial. 
Without focusing on specific tokens, Self-Familiarity~\cite{SelfFAMILIARITY} estimates an LLM's familiarity with the requested facts through multi-round conversations with the LLM, requiring carefully crafted prompts to engage the LLM multiple times, resulting in low prediction efficiency.
Alternatively, researchers fully fine-tune the LLM for NFP~\cite{LLMKnowWhatTheyKnow} (termed Self-Evaluation).
However, this approach incurs significant computational costs and may hinder the LLM's generalization ability~\cite{GeneralizationOfFinetunedLLMs}.
In contrast, \smodel exhibits good applicability, high efficiency, and enables cross-LLM domain adaptation.
\footnote{Hidden question representation has been utilized for predicting an LLM's self-consistency~\cite{ProbingNFP}. Notably, self-consistency does not equate to factuality; for instance, an LLM can consistently produce incorrect answers.}

\section{Preliminary}\label{sec:pre}
\subsection{Problem Definition}

\begin{definition}
    \textbf{Non-Factual Response.}
    Given an LLM $m \in \mathcal{M}$ and a fact-seeking question $q \in \mathcal{Q}$, $m$ generates an answer $s$. If the answer $s$ fails to convey the queried fact, it is a non-factual response.
\end{definition}

\begin{problem}
	\textbf{Non-Factuality Prediction in an LLM (NFP).} Given an LLM $m \in \mathcal{M}$ and a fact-seeking question $q \in \mathcal{Q}$, the objective is to learn a function $f\left(m, q\right) \rightarrow y$, where $y=1$ if $m$ will generate a non-factual response to $q$ and $y=0$ otherwise.
\end{problem}

\begin{problem}
	\textbf{Transferable Cross-LLM NFP.}
 Given LLMs $m_{1}, m_{2} \in \mathcal{M}$ and a fact-seeking question set $\mathcal{Q}$, NFP labels have been constructed based on $\mathcal{Q}_{train} \subset \mathcal{Q}$ for $m_{1}$, deriving a training set $\{\left(\left(m_{1}, q_i\right), y_{1,i}\right)\}_{q_i\in \mathcal{Q}_{train}}$. The goal is to utilize the training set and $m_2$ to learn a function $f\left(m, q\right) \rightarrow y$, where $m\in \{{m_{1}, m_{2}}\}$ and $q \in \mathcal{Q}$.
\end{problem}

\subsection{NFP Datasets}\label{sec:NFP construction}
\vpara{Dataset Construction.}
Given an LLM $m$ and a fact-seeking QA dataset, for each question $q \in \mathcal{Q}$, we assign a binary label $y$ to the $\left(m, q\right)$ pair, where $y=1$ if $m$ fails to generate the golden answer for $q$, and $y=0$ otherwise. Our goal is to predict the labels prior to answer generation.
Notably, a fact-seeking question asks for objective and verifiable information, such as dates, locations, and entities. Examples include ``In which year was the Eiffel Tower built?'' and ``Which city is the capital of France?''. Their answers are naturally short and precise.
Therefore, we follow previous work~\citep{whenToUseExternalKnowledge} to mark an LLM's response as non-factual (i.e., $y=1$) if no sub-string of the response matches any of the gold answers.\footnote{The labeling method ensures accurate labels of all positive samples. We randomly sample 20 negative samples from each NFP dataset, deriving $20\times12=240$ negative samples, and manually checked their labels' quality. Given that all positive samples constitute 72.2\% of the dataset, the ratio of correct labels is 97.0\%.}
We consider four LLMs and three QA datasets in the main body of the paper, deriving $4\times3=12$ NFP datasets.
In each NFP dataset, consisting of samples in the form of $\left((m, q), y\right)$, we randomly sample 20\% data for training, 10\% data for validation, and use the remaining data for testing. 

\vpara{LLMs \& QA Datasets.} LLaMA2-7B-Chat~\citep{llama2}, LLaMA3-8B-Instruct~\citep{llama3}, Mistral-7B-Instruct-v0.2~\citep{mistral}, and Qwen2-1.5B-Instruct~\citep{qwen2} are used for experiments. These LLMs have been instruction-tuned for conversational engagement.
We pose questions from three widely-used QA datasets: PopQA (PQ)~\citep{whenToUseExternalKnowledge}, Entity Questions (EQ)~\citep{sciavolino-etal-2021-simple}, and Natural Questions (NQ)~\citep{naturalquestions}.
Detailed statistics of these datasets are provided in Appendix~\ref{appendix:dataset}.
To ensure reproducibility, we set each LLM's decoding strategy to greedy search rather than top-$p$ or top-$k$ sampling. We have also run the sampling-based decoding, and find that the paper's experimental conclusions still hold true.

\section{Methodology}
\subsection{\model}
Given an LLM $m$ and a fact-seeking question $q$, we can quickly acquire the hidden states corresponding to input tokens.
In a certain layer, we use the hidden states corresponding to the last input token as the question's hidden representation $\mathbf{x}$.
Then we use an encoder $g_{{enc}}$ to transform the question's hidden representation into a latent feature space, where we presume that the NFP patterns are represented.
Afterwards, a linear classifier $g_{{clf}}$ is set upon $g_{{enc}}$ for classification. 
Formally, based on the $\ell$-th hidden layer of $m$, 
\smodel predicts by, 
\begin{equation}
    \mathbf{p} = g_{{clf}}\left( g_{{enc}}\left(m_{\leq \ell}\left(q\right)\right)\right) = g_{{clf}}\left( g_{{enc}}\left(\mathbf{x}\right)\right)
\end{equation}
\noindent where $m_{\leq \ell}\left(\cdot\right)$ denotes the function composed of the $\ell$-th transformer layer and its preceding layers, $g_{{enc}}$ is implemented by a lightweight multi-layer perceptron (MLP)\footnote{Our goal is to verify that hidden question representations contain useful patterns for the NFP task. Exploring other model architectures for the NFP pattern extraction is beyond the scope of this paper.}, $g_{{clf}}$ is implemented by a linear layer with the Softmax function, and $\mathbf{p}$ is a two-dimensional vector revealing the probability of (not) producing non-factual responses. 
Based on a set of labeled NFP instances $\{\left(m_{\leq \ell}\left(q_i\right), y_i\right)\}_{q_i \in \mathcal{Q}_{train} \cup \mathcal{Q}_{val}}$, where $Q_{train}$ and $Q_{val}$ denote question sets used for training and validation, respectively, we can train a \smodel for $m$ with the classic cross-entropy (CE) loss.

\begin{figure}
	\centering
		\includegraphics[width=3in]{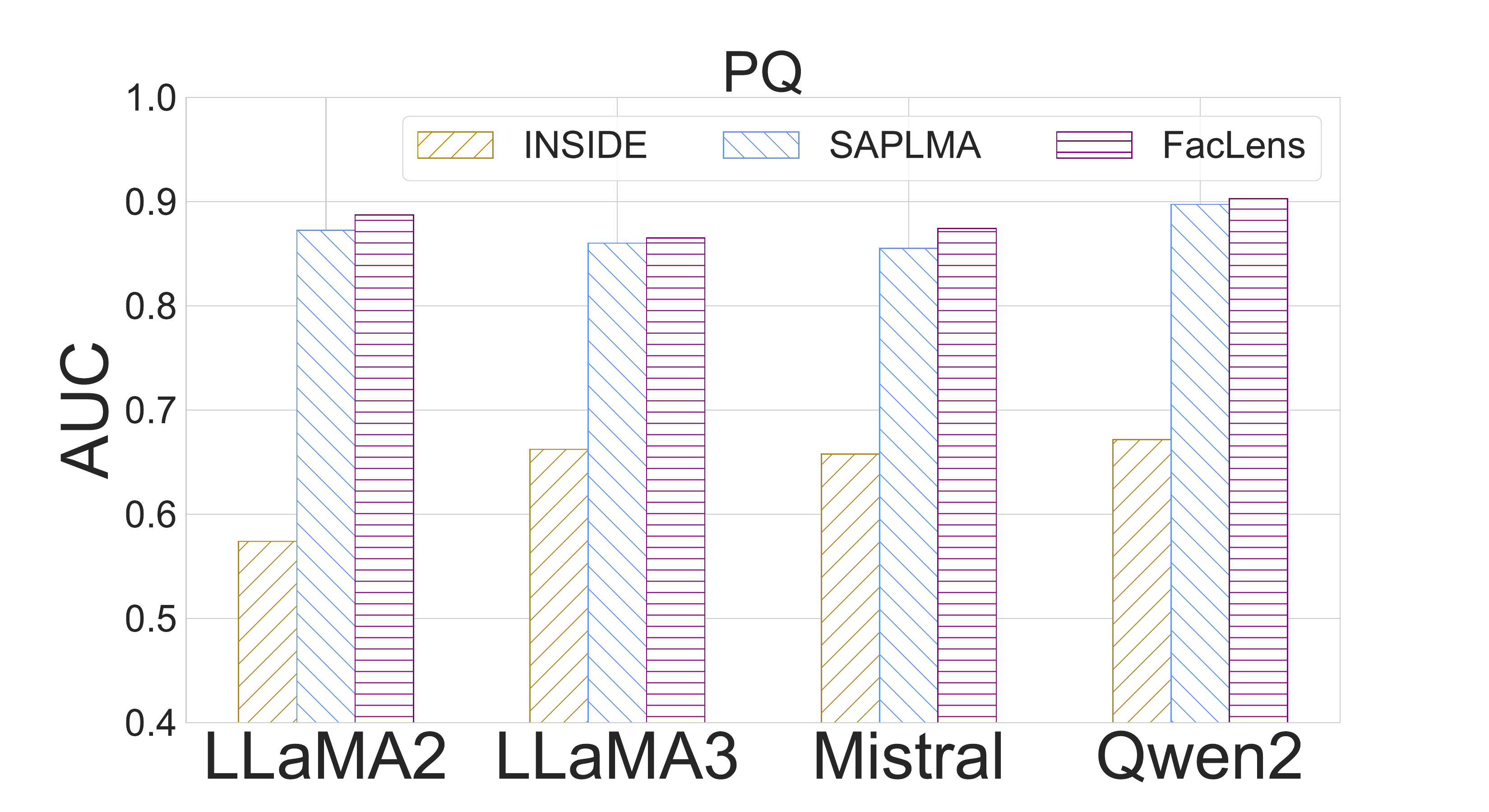}
	   \caption{Performance comparison of \smodel (ante-hoc NFP) with INSIDE and SAPLMA (post-hoc NFD). Trends are consistent on EQ and NQ (see Appendix~\ref{appendix: Ante-Hoc NFP vs. Post-Hoc NFD}).}
    \label{fig:compare_with_post_hoc_NFD}
\end{figure}

\textbf{\textit{Observation}: While post-hoc NFD methods leverage more information, the ante-hoc \smodel has the potential to perform better.}\label{appendix:appendix_comparison_with_NFD_methods}
In Figure~\ref{fig:compare_with_post_hoc_NFD}, we compare \smodel with SAPLMA~\citep{SAPLMA} and INSIDE~\citep{INSIDE}, two representative post-hoc methods that leverage hidden representations of LLM-generated responses to identify non-factual responses.
While post-hoc methods leverage more information (i.e., the LLM-generated responses), \smodel shows comparable and even better performance.

\label{sec:faclens}
\subsection{Transferability of \model}\label{sec: pilot study transferability of faclens}
When it comes to multiple LLMs, the construction of NFP training data becomes resource-intensive and time-consuming, because each LLM needs to conduct costly response generation (see Section~\ref{sec:NFP construction}). 
Fortunately, we discover the transferability of \model, which allows us to label training data for just one LLM and adapt the \smodel to support other LLMs.
In Appendix~\ref{appendix: Comparison Between Different Labeling Methods}, we illustrate the more efficient process of training data construction enabled by the transferability of \model.

\vpara{Why Domain Adaptation is Effective for Transferring \smodel Across LLMs.}
Domain adaptation (DA) is an approach in transfer learning that transfers information from a source domain to improve performance in a target domain~\citep{DomainAdaptationFeatureRrepresentations,unsupervisedDomainAdaptation,conceptshift}.
The premise of DA is that the source and target domains have distinct marginal probability distributions $P(\mathbf{X})$, but share similar conditional probability distributions $P(\mathbf{y}|\mathbf{X})$ (i.e., no significant concept shift)~\citep{unsupervisedDomainAdaptation,conceptshift}.
Here we refer to the domain as,
\begin{remark}
Let the variable $\mathbf{X}$ represent the hidden question representation in an LLM. A data domain $D$ refers to a collection of hidden question representations sourced from a certain LLM. \end{remark}

Different domains naturally have different $P(\mathbf{X})$.
If $P(\mathbf{y}|\mathbf{X})$ of different domains exhibit similar forms, we can perform DA to apply \smodel to other LLMs without new labels for training.

Now we verify that different data domains indeed have similar conditional distributions $P(\mathbf{y}|\mathbf{X})$ by introducing a mixture domain $D_{mix}$, whose joint probability distribution is,
\begin{equation}
\begin{split}
    P_{mix}(\mathbf{X}, \mathbf{y}) &= \sum\nolimits_{i=1}^{M} \alpha_{i} \cdot P_{m_i}(\mathbf{X}, \mathbf{y}) \\
    &\quad \text{s.t.} \quad \sum\nolimits_{i=1}^{M} \alpha_{i} = 1
\end{split}
\end{equation}
where $M$ is the number of individual data domains (i.e. the number of different LLMs), $m_i$ denotes the $i$-th LLM, and $0 < \alpha_i < 1$ represents the proportion of $D_i$ in the mixture domain. Here we set $\alpha_i = \frac{1}{M}$.

Therefore, the \smodel trained on the mixture domain follows the conditional distribution,
\begin{equation}
\begin{split}
    P_{mix}(\mathbf{y}|\mathbf{X}) &= \sum\nolimits_{i=1}^{M} \beta_{i}(\mathbf{X}) \cdot P_{m_i}(\mathbf{y}|\mathbf{X}), \\
    \beta_{i}(\mathbf{X}) &= \frac{\alpha_i\cdot P_{m_i}(\mathbf{X})}{\sum_{j=1}^{M}\alpha_j\cdot P_{m_j}(\mathbf{X})}
\end{split}
\end{equation}
It is readily derived that $\sum\nolimits_{i=1}^{M} \beta_{i}(\mathbf{X}) = 1$, and $0 < \beta_i(\mathbf{X}) < 1$ if $P_{m_1}(\mathbf{X}), P_{m_2}(\mathbf{X}), \cdots, P_{m_M}(\mathbf{X})$ are not disjoint. 
If there are no concept shifts between
individual data domains, we have,
\begin{equation}
\begin{split}
    P_{mix}\left(\mathbf{y}|\mathbf{X}\right) &= P_{m_1}\left(\mathbf{y}|\mathbf{X}\right) = P_{m_2}\left(\mathbf{y}|\mathbf{X}\right) \\
    &= \cdots = P_{m_M}\left(\mathbf{y}|\mathbf{X}\right)
\end{split}
\label{eq:no_concept_shift}
\end{equation}
Conversely, if significant concept shifts exist between individual domains,  Eq.~\ref{eq:no_concept_shift} is not valid, as there must exist at least a domain $D_i$ where $P_{mix}\left(\mathbf{y}|\mathbf{X}\right)$ is very different from $P_{m_i}\left(\mathbf{y}|\mathbf{X}\right)$. Consequently, on the test set of domain $D_i$, $f_{mix}$ will noticeably underperform $f_{m_i}$, where $f_{mix}$ is trained in $D_{mix}$, and $f_{m_i}$ is trained in $D_i$.
For simplicity, we use $f_{m}$ to denote a \smodel trained on an individual domain.

\begin{figure}
	\centering
		\includegraphics[width=3in]{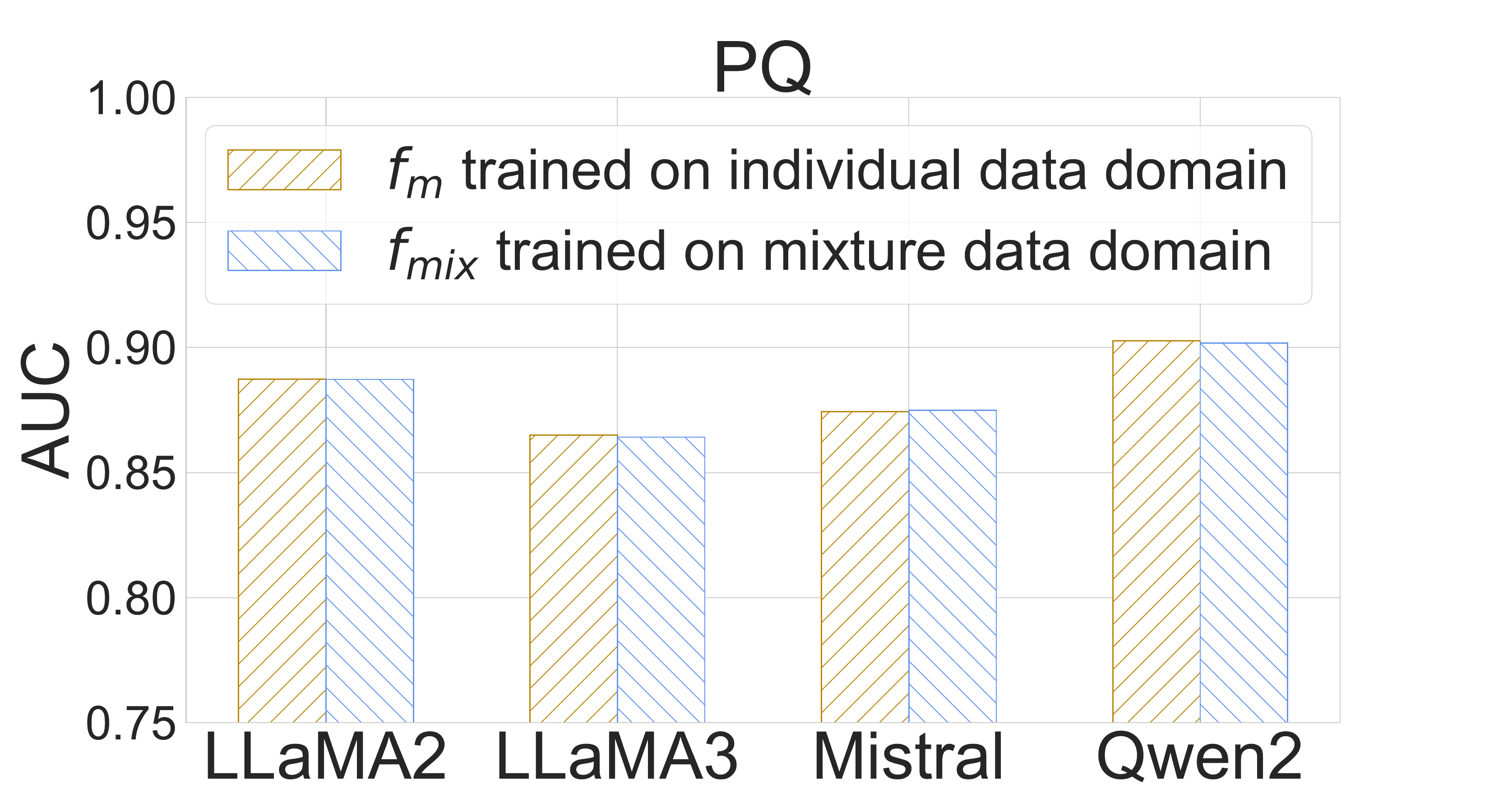}
	   \caption{Performance comparison between $f_{m}$ and $f_{mix}$. Similar performance suggests no significant concept shift across different domains. Trends are consistent on EQ and NQ (See Appendix~\ref{appendix: Demonstration of FacLens's Transferability})}
    \label{fig:pilot_study_for_DA}
\end{figure}

\begin{figure}
	\centering
		\includegraphics[width=1.45in]{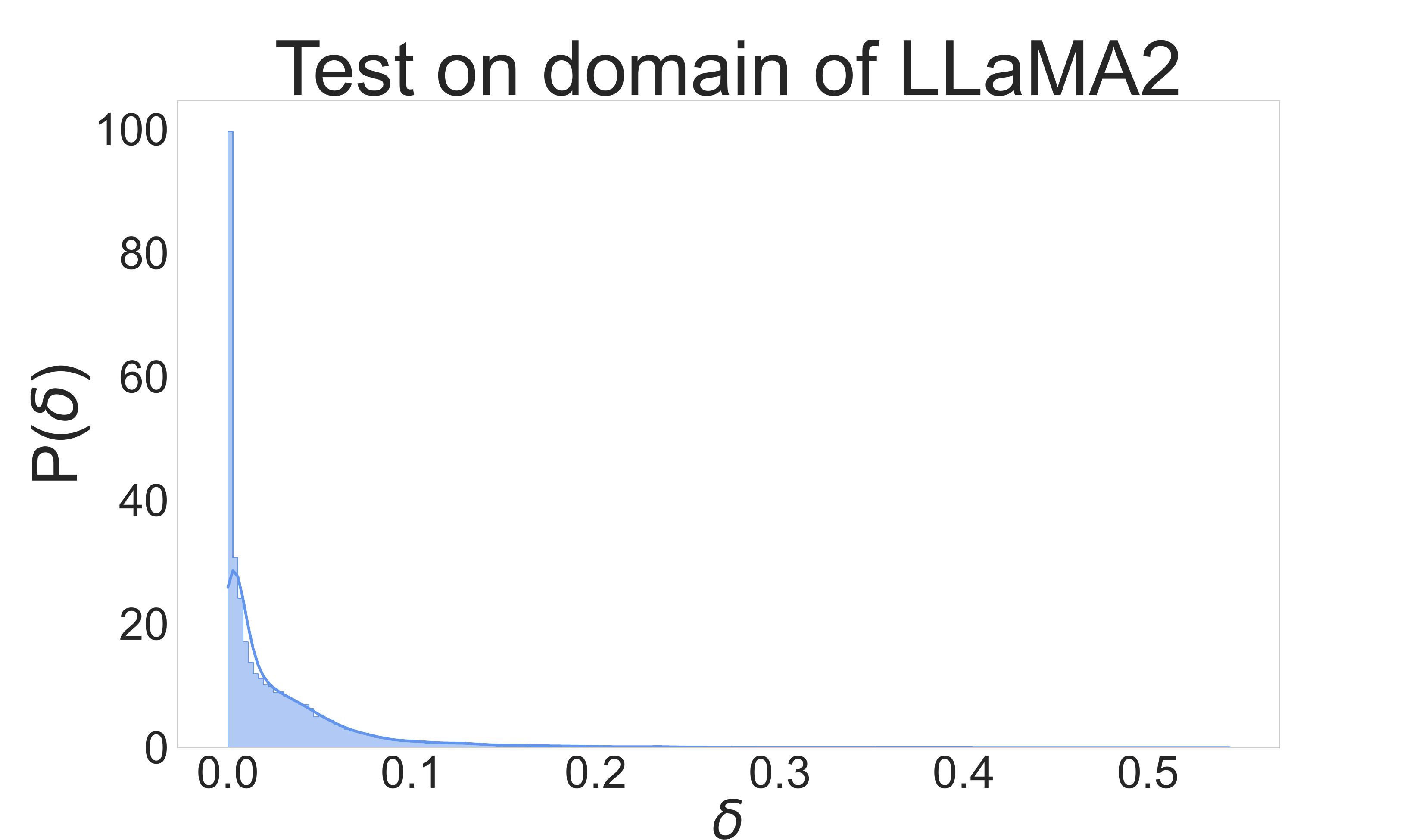}
            \includegraphics[width=1.45in]{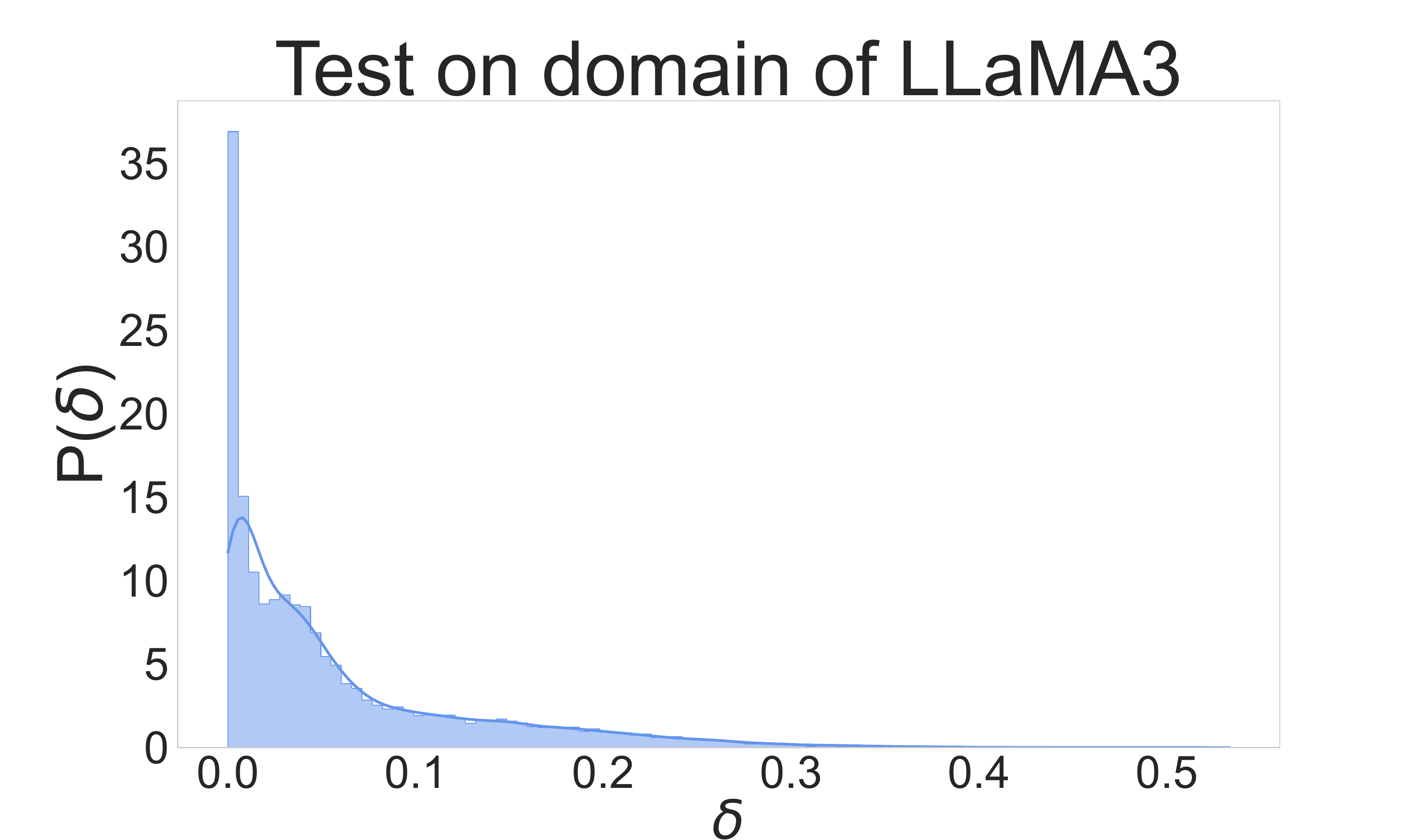}
          \includegraphics[width=1.45in]{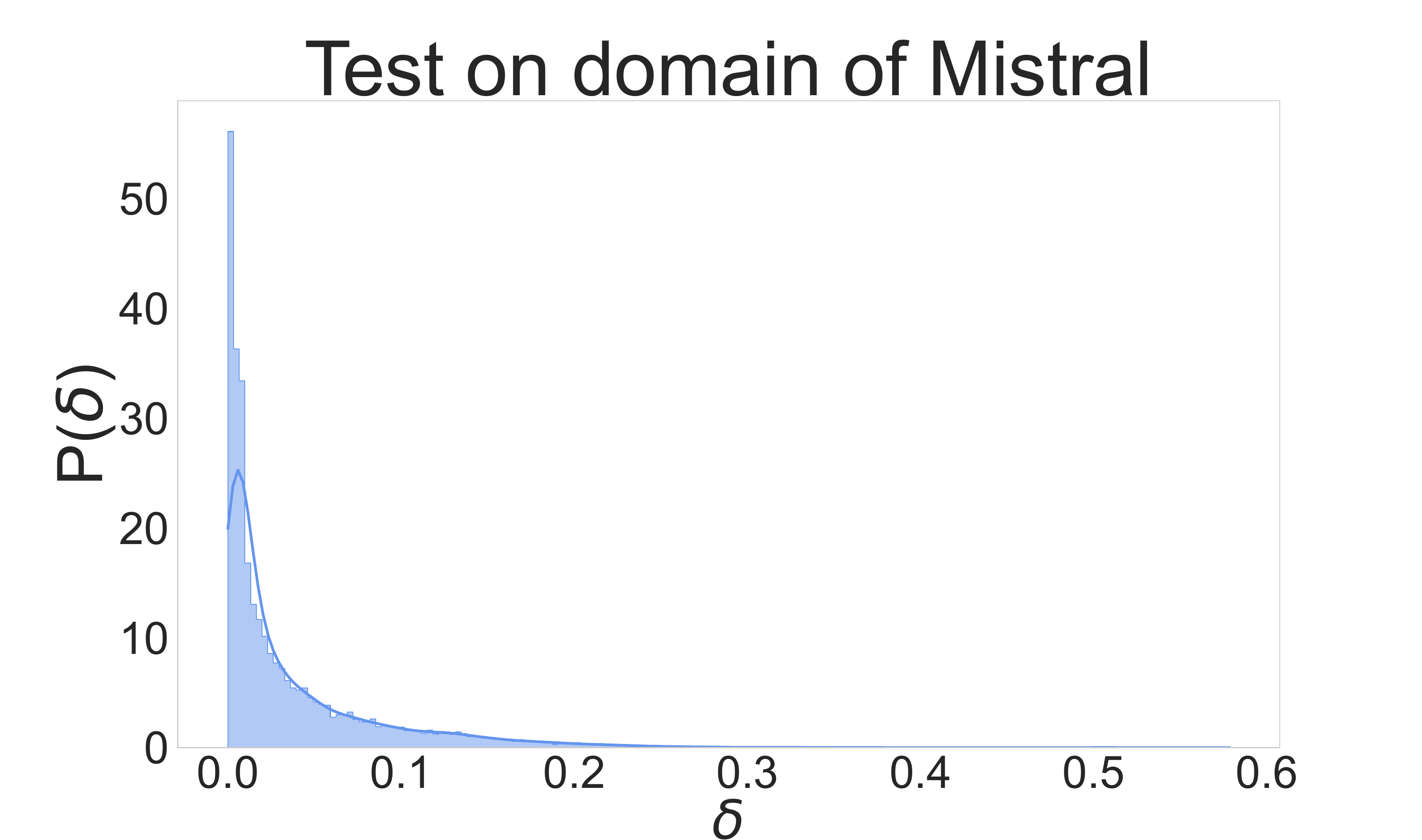}          
          \includegraphics[width=1.45in]{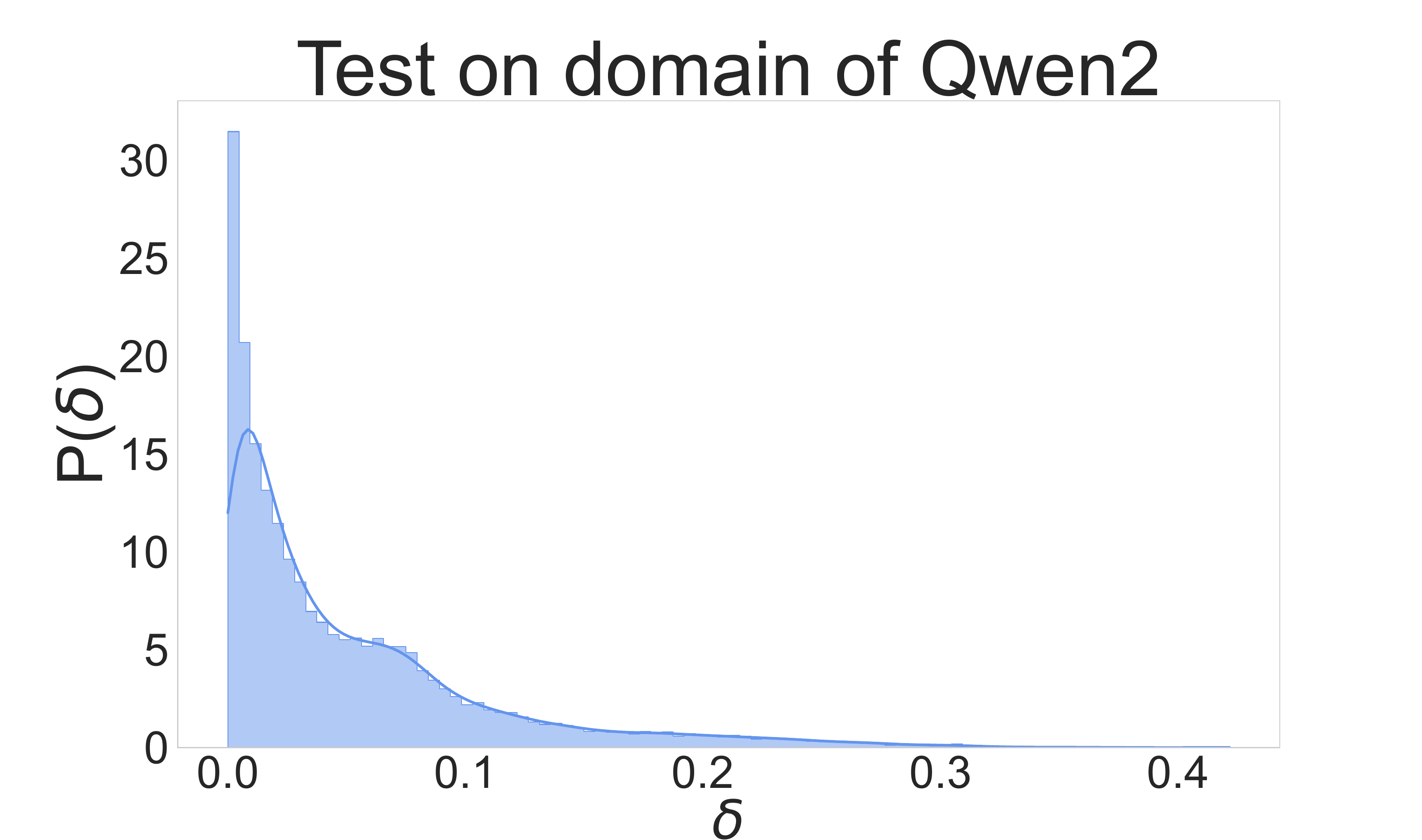}
	   \caption{Distribution of $\delta$ between an individual-domain \smodel and the mixture-domain \smodel overall questions (questions from PQ, EQ, and NQ).}
    \label{fig:pilot_study_for_DA_density}
\end{figure}

\begin{figure}[ht]
	\centering
		\includegraphics[width=2.5in]{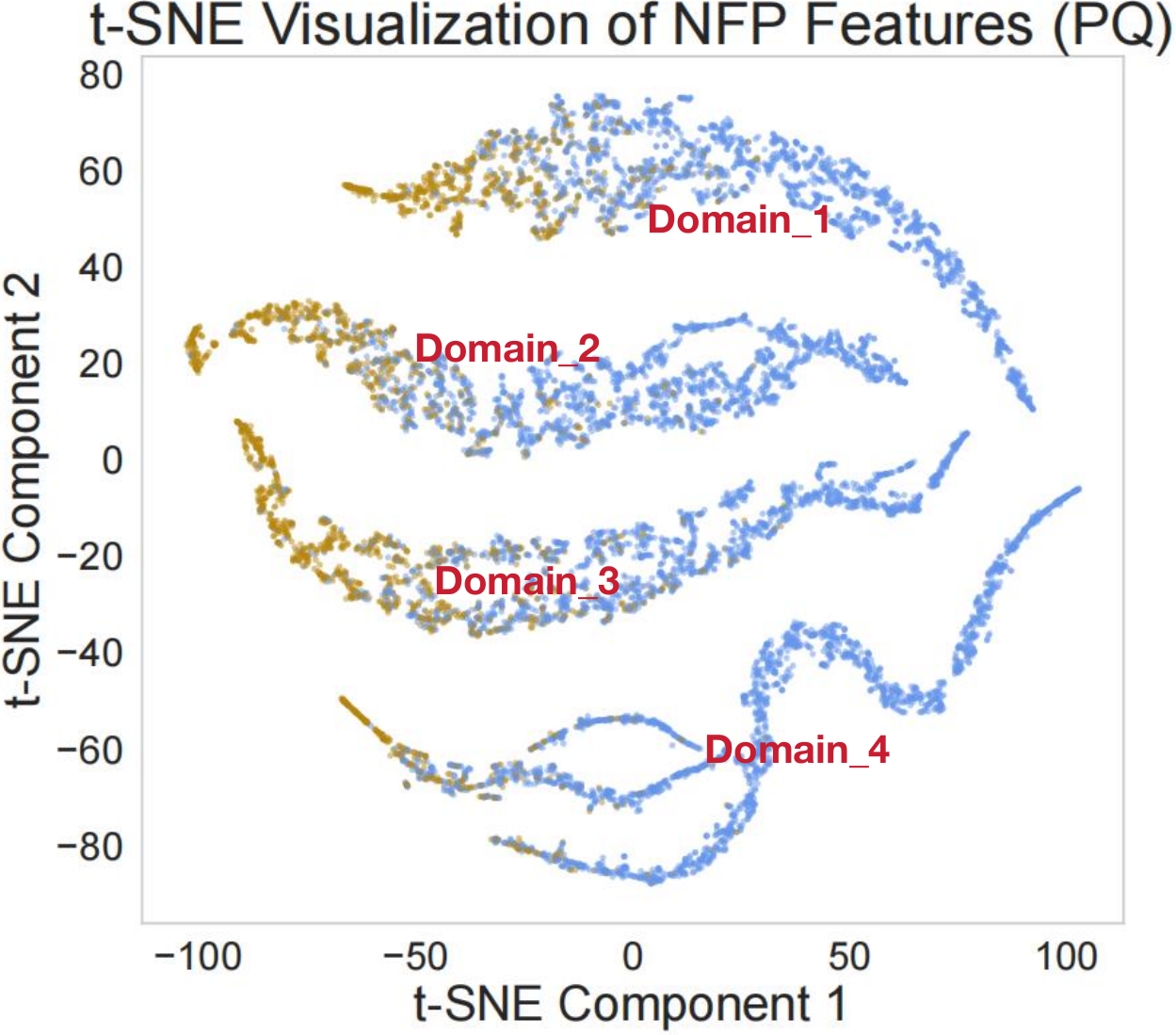}
	   \caption{Visualization of NFP features extracted from different LLMs' hidden question representations, where each domain corresponds to an LLM, and the blue and dark yellow points denote the positive and negative samples, respectively. Trends are consistent on EQ and
NQ (see Appendix~\ref{appendix: Demonstration of FacLens's Transferability}).}
    \label{fig:pilot_study_for_DA_visualization}
\end{figure}

Given a specific fact-seeking QA dataset, we have four individual domains as our experiments consider four LLMs. Each individual domain has its training, validation, and test sets. The training sets of all the individual domains form the training set of the mixture domain. Notably, the hidden dimension of Qwen2-1.5B-Instruct is different from that of the other three LLMs, so we introduce an additional linear layer to reshape Qwen2's hidden question representations to match the dimension of the other three LLMs.

\textit{\textbf{Observation 1.}} 
Figure~\ref{fig:pilot_study_for_DA} shows the results on the PQ dataset, and we can observe that $f_{mix}$ exhibits comparable performance to $f_{m}$ on the test set of the corresponding individual domain, indicating similar $P(\mathbf{y}|\mathbf{X})$ across different domains. Hence, \textbf{we can conduct DA to derive a \smodel used by different LLMs}.

\textit{\textbf{Observation 2.}} 
We measure concept shift between an individual domain and the mixture domain by $\delta = || \mathbf{p}^{m}(y=1|\mathbf{x}) - \mathbf{p}^{mix}(y=1|\mathbf{x})||$, where $\mathbf{p}^{m}$ is computed by $f_{m}$ and $\mathbf{p}^{mix}$ by $f_{mix}$. The values of $\delta$ are mostly near zero (see Figure~\ref{fig:pilot_study_for_DA_density}), indicating that $P_{m_1}(\mathbf{y|X}), P_{m_2}(\mathbf{y|X}), \dots, P_{m_M}(\mathbf{y|X})$ and $P_{mix}(\mathbf{y|X})$ have similar distributions.

\textit{\textbf{Observation 3.}}
Given hidden question representations from different LLMs, we use the encoder of $f_{mix}$ to extract the NFP features and visualize them with t-SNE. The results of PQ are shown in Figure~\ref{fig:pilot_study_for_DA_visualization}, where the positive and negative samples are represented by points of different colors.
Although these points are sourced from different LLMs, we can see that a unified classification boundary can be applied to them, further demonstrating the similar $P(\mathbf{y}|\mathbf{X})$ across different domains.

\subsection{Cross-LLM \model}\label{sec: cross-llm faclens}

We have verified that we can use DA to train a cross-LLM \smodel for LLM $m_j$ leveraging label information from LLM $m_i$.
Here the distribution shift between a source domain $D_{S}$ and a target domain $D_{T}$, is due to the difference of LLMs. 
$D_S$ has labeled data, yet $D_T$ has no label information.

\vpara{Unsupervised Domain Adaptation.}
The classic Maximum Mean Discrepancy (MMD) loss~\citep{MMD} is used to find a domain-invariant NFP feature space, based on which \smodel predicts the labels.
The MMD loss calculates the distance between two distributions in the reproducing kernel Hilbert space (RKHS)~\citep{RKHSforDistributions}.
We denote the NFP features in the source and target domains as $Z_{S} = \{\mathbf{z}_{S, i}\}_{i=1}^{N_S}$ and $Z_{T} = \{\mathbf{z}_{T, j}\}_{j=1}^{N_T}$, respectively, where $\mathbf{z}_{S, i} = g_{enc}\left(\mathbf{x}_{S, i}\right)$, and $\mathbf{z}_{T, j} = g_{enc}\left(\mathbf{x}_{T, j}\right)$.
The encoder $g_{enc}$ in \smodel is optimized by minimizing the MMD loss,
\begin{equation}
\begin{split}
    \mathcal{L}_{\text{MMD}}\left(Z_{S}, Z_{T}\right) &= \frac{1}{N_S^2}\sum_{i, j=1}^{N_S} k(\mathbf{z}_{S, i}, \mathbf{z}_{S, j}) \\
    &\quad + \frac{1}{N_T^2}\sum_{i, j=1}^{N_T} k(\mathbf{z}_{T, i}, \mathbf{z}_{T, j}) \\
    &\quad - \frac{2}{N_SN_T}\sum_{i=1}^{N_S}\sum_{j=1}^{N_T} k(\mathbf{z}_{S, i}, \mathbf{z}_{T, j})
\end{split}
\end{equation}
\noindent where $N_S = N_T = |\mathcal{Q}_{train}|$ is the number of questions for training, and $k\left(\cdot\right)$ denotes a kernel function. We extract hidden question representations from the LLM's middle layer.

Importantly, we also use the CE loss to optimize $g_{enc}$ and $g_{clf}$ with the labeled data in $D_S$, which collaborates the MMD loss to find the latent feature space for NFP. Finally, the loss function is,
\begin{equation}
    \mathcal{L}_{\text{DA}} = \mathcal{L}_{\text{MMD}} 
    + \frac{1}{N_S} \sum_{i=1}^{N_S}\mathcal{L}_{\text{CE}}\left(g_{clf}\left(\mathbf{z}_{S, i}\right), y_{S, i}\right)
\label{eq:DA}
\end{equation}
Notably, if LLMs have distinct hidden dimensions, we introduce an additional linear layer to reshape the target domain's hidden question representations to match the dimension of the source domain's hidden question representations. We demonstrate that \smodel can transfer across LLMs of distinct hidden dimensions in Figure~\ref{fig:transfer_performance_linear_kernel} and Appendix~\ref{appendix: cross-LLM faclens different dimension}.
Besides, we discuss the choice of kernel function for MMD loss in Appendix~\ref{appendix:kernel selection}. 

\begin{table*}[ht]
   \scriptsize
   \centering
   \renewcommand\arraystretch{1}
   \begin{threeparttable}
    \newcolumntype{?}{!{\vrule width 1pt}}
    \setlength{\tabcolsep}{1.8mm}
    \begin{tabular}{l?ccc?ccc?ccc?ccc}
        \toprule
        &\multicolumn{3}{c?}{\textbf{LLaMA2}}&\multicolumn{3}{c?}{\textbf{LLaMA3}}&\multicolumn{3}{c?}{\textbf{Mistral}} & \multicolumn{3}{c}{\textbf{Qwen2}}\\
        \cmidrule{2-13}
        & \textbf{PQ}& \textbf{EQ} & \textbf{NQ}  & \textbf{PQ}& \textbf{EQ} & \textbf{NQ} 
 & \textbf{PQ}& \textbf{EQ} & \textbf{NQ} & \textbf{PQ}& \textbf{EQ} & \textbf{NQ} \\
        \midrule
         PPL & 72.5 & 67.1 & 56.4 & 69.8 & 65.5 & 53.9 & 69.1 & 67.2 & 57.7 & 74.1 & 64.6 & 57.4\\
         Prompting & 72.7 & 67.8 & 58.1 & 70.6 & 64.9 & 57.2 & 72.2 & 66.0 & 65.5 & 73.0 & 74.7 & 57.1\\
         Entity-Popularity& 79.0 & -- & -- & 75.9 & -- & -- &  77.6 & -- & -- & 67.9 & -- & --\\
         SAT Probe & 85.1 & 79.3 & -- & 83.4 & 81.5 & -- & 84.4 & 81.9 & -- & 88.5 & 81.9 & --\\
         Self-Familiarity & 59.1 & 64.9 & 55.8 & 61.8 & 68.4 & 52.0 & 57.1 & 64.9 & 54.2 & 54.1 & 61.8 & 57.6 \\
         LoRA (Parameter-Efficient FT)& 88.2 & 84.8 & 67.0 & 86.1 & 83.8 & 63.2 & 84.1 & 81.8 & 65.7 & 90.0 & 85.1 & 73.5\\
         Self-Evaluation (Fully FT)& 88.5 & 85.2 & 68.8 & 85.7 & \textbf{85.8} & 63.9 & 83.5 & 80.9 & 61.9 & 89.7 & \textbf{86.6} & 71.3\\
         
        \midrule
        \model-ent (avg, last layer) & 76.0 & 79.6 & 60.4 & 75.8 & 77.7 & 57.4 & 76.8 & 77.8 & 59.2 & 84.6 & 77.7 & 65.2\\
        \model-ent (avg, 2$^{\text{nd}}$ to last layer) & 77.9 & 80.5 & 60.4 & 76.2 & 79.0 & 58.0 & 77.1 & 78.3 & 60.5 & 84.5 & 78.6 & 65.1\\
        \model-ent (avg, middle layer) & 81.7 & 81.2 & 60.6 & 79.2 & 81.0 & 58.6 & 81.4 & 82.4 & 61.5 & 87.0 & 82.2 & 65.4\\
        
         \midrule
        \model-ent (last token, last layer) & 81.4 & 81.7 & 60.6 & 78.9 & 79.6 & 55.3 & 80.9 & 80.9 & 59.3 & 87.4 & 81.7 & 64.4\\
        \model-ent (last token, 2$^{\text{nd}}$ to last layer) & 82.3 & 82.1 & 60.1 & 78.1 & 79.7 & 57.8 & 81.6 & 81.9 & 59.7 & 87.6 & 81.7 & 63.9 \\
        \model-ent (last token, middle layer) & 83.5 & 81.4 & 61.2 & 79.9 & 81.0 & 60.0 & 82.9 & 82.8 & 60.5 & 88.0 & 81.5 & 63.5 \\
        \midrule
        
        \smodel (last token, last layer) & 88.7 & 84.9 & 69.1 & 86.1 & 84.1 & 64.7 & 86.1 & 84.4 & 71.7 & 90.0 & 85.9 & \textbf{74.0}\\
        \smodel (last token, 2$^{\text{nd}}$ to last layer) & \textbf{88.8} & 85.0 & 67.7 & 86.1 & 84.1 & 65.6 & 87.0 & \textbf{85.7} & \textbf{72.1} & \textbf{90.7} & 85.6 & 72.4\\
        \smodel (last token, middle layer) & 88.7 & \textbf{85.6} & \textbf{69.5} &\textbf{86.5} & 85.0 & \textbf{68.9} & \textbf{87.4} & 85.4 & 71.4 & 90.3 & 86.4 & 71.6\\
        \bottomrule
    \end{tabular}
    \small
    \begin{tablenotes}
        \item ``--'' means the method is not suitable for the QA dataset. We give the detailed explanation in the appendix~\ref{appendix: baseline not suitable for some datastes}.
        ``avg'' refers to the averaged hidden representation of the input entities' tokens or a question's tokens. ``last token'' refers to the hidden representation of the last token in the input entities or the question.
        The question consists of a chat template and the original question, where the chat template can prompt the LLM to better respond. Due to space limitation, we show the performance of \smodel (avg) in Appendix~\ref{appendix:FacLens(avg)}.
    \end{tablenotes}
    \end{threeparttable}
    \caption{Prediction performance of different NFP methods (AUC \%).}
    \label{tab:exp_overall_evaluation}
\end{table*}

\vpara{Question-Aligned Mini-Batch Training.}
In order to address GPU out-of-memory issues, \smodel employs mini-batch training for DA. In each mini-batch, we sample two question sets, $\mathcal{\overline{Q}}_{S}$ and $\mathcal{\overline{Q}}_{T}$, from $Q_{train}$, for two domains.
This raises a question: are $\mathcal{\overline{Q}}_{S}$ and $\mathcal{\overline{Q}}_{T}$ identical?
Given a range of questions, the distribution $P(\mathbf{Z})$ should be determined by the LLM.
In a mini-batch, the number of sampled questions is limited, so the estimation of $P_{S}\left(\mathbf{Z}\right)$ and $P_{T}\left(\mathbf{Z}\right)$ within the mini-batch is likely to be affected by the sampling process. Hence, we propose to use the same question set for two domains in each mini-batch, i.e., $\mathcal{\overline{Q}}_{S} = \mathcal{\overline{Q}}_{T}$, to alleviate the influence of sampling process in estimating the true distance between $P_{S}\left(\mathbf{Z}\right)$ and $P_{T}\left(\mathbf{Z}\right)$.

\section{Experiments}
\subsection{Experimental Setup}
We compare \smodel with existing NFP methods, which have been introduced in Section~\ref{sec:related}, including \textbf{Entity-Popularity}~\citep{whenToUseExternalKnowledge}, \textbf{SAT Probe}~\citep{SATProbe}, \textbf{Self-Familiarity}~\citep{SelfFAMILIARITY}, and \textbf{Self-Evaluation}~\citep{LLMKnowWhatTheyKnow}.
As Self-Evaluation fully fine-tunes the LLM for NFP, we adopt \textbf{LoRA}~\citep{LoRA} as an additional baseline to conduct parameter-efficient fine-tuning.
We also consider a \textbf{Prompting}-based method, which directly asks the LLM whether the LLM knows the factual answer to the given question.
Moreover, inspired by using the perplexity to evaluate the factual precision of responses~\citep{FActScore}, we consider \textbf{perplexity (PPL)} on the input question as a baseline (see Appendix~\ref{appendix:form_of_PPL}).
Due to space limitation, we provide the hyper-parameter settings in Appendix~\ref{appendix:baseline_experimental_setting}.
As the number of positive samples is larger than that of negative samples (see Table~\ref{tab:response accuracy}), we adopt AUC, a common metric for imbalanced binary classification, as the evaluation metric.

\begin{table*}[ht]
   \scriptsize
   \centering
   \renewcommand\arraystretch{1}
   \setlength{\tabcolsep}{1.7mm}
    \begin{threeparttable}
    \newcolumntype{?}{!{\vrule width 1pt}}
    \begin{tabular}{l?c?c?c?c}
        \toprule
        & \textbf{Training-Free} &\textbf{Transferable}
 &\makecell[c]{\textbf{Training Time} \\ \textbf{Per Epoch (avg)}} & \makecell[c]{\textbf{Prediction Time} \\ \textbf{Per Question (avg)}} \\
        \midrule
        Self-Familiarity & Yes & -- & -- & 5.838s\\
        Prompting & Yes  & -- & -- & 0.115s\\
        PPL & Yes & -- & -- & 0.044s\\
        LoRA (Parameter-Efficient FT)& No & No & 116.500s (1 * 80G A800) & 0.038s\\
        Self-Evaluation (Fully FT)& No & No & 184.778s (4 * 80G A800)& 0.028s\\
        SAT Probe & No & No &  0.010s (1 * 80G A800) & 0.037s\\
        \model & No & Yes & 0.012s (1 * 80G A800)& 0.016s\\
        \bottomrule
    \end{tabular}
    \small
    \begin{tablenotes}
        \item Note: Here \smodel denotes \smodel (last token, middle layer). 
        SAT Probe involves a feature extraction step, where attention weights are extracted for probing, taking 132.860s.
        Before training \model, we extract hidden question representations, a process that takes only 71.856s. 
    \end{tablenotes}
    \end{threeparttable}
    \caption{Efficiency evaluation of \smodel (seconds).}
    \label{tab:efficacy}
\end{table*}

\subsection{Experimental Results}

\textbf{Probing hidden question representations for NFP in fact-seeking QA yields promising results.}
As reported in Table~\ref{tab:exp_overall_evaluation},
\smodel exhibits promising performance compared to the baselines, and it favors hidden representations from LLMs' middle layers.
Specifically, Self-Evaluation and LoRA can be regarded as two special representation-based NFP methods. While both perform well, \model, with much fewer trainable parameters, offers higher efficiency and yields results comparable to, or even surpassing, those of Self-Evaluation and LoRA.
Compared to SAT Probe, \smodel demonstrates that hidden representation could be more useful than attention weights for NFP.
Furthermore, we conduct main experiments on HotpotQA~\cite{hotpotqa}, a more complex multi-hop QA dataset, in Appendix~\ref{appendix: Evaluation on the More Complex QA Dataset HotpotQA}. The results show that \smodel continues to perform well.
We also developed a demo to collect more questions and provide the case study in Appendix~\ref{appendix:case_study}.

\textbf{Question-level modeling is more effective than focusing on specific tokens.}
Inspired by entity-centric studies, we propose \model-ent, which feeds the representations of input entity tokens into \model.\footnote{Entities, e.g., persons, locations, organizations, are identified using the Stanza NLP Package~\citep{Stanza}.} As shown in Table~\ref{tab:exp_overall_evaluation}, \smodel consistently surpasses \model-ent, Entity-Popularity, Self-Familiarity, and SAT Probe, suggesting that overemphasizing specific tokens of the input question may mislead the predictions.

\begin{figure*}[ht]
    \centering
    \subfigure[AUC Results on PQ]{\includegraphics[width=0.25\textwidth]{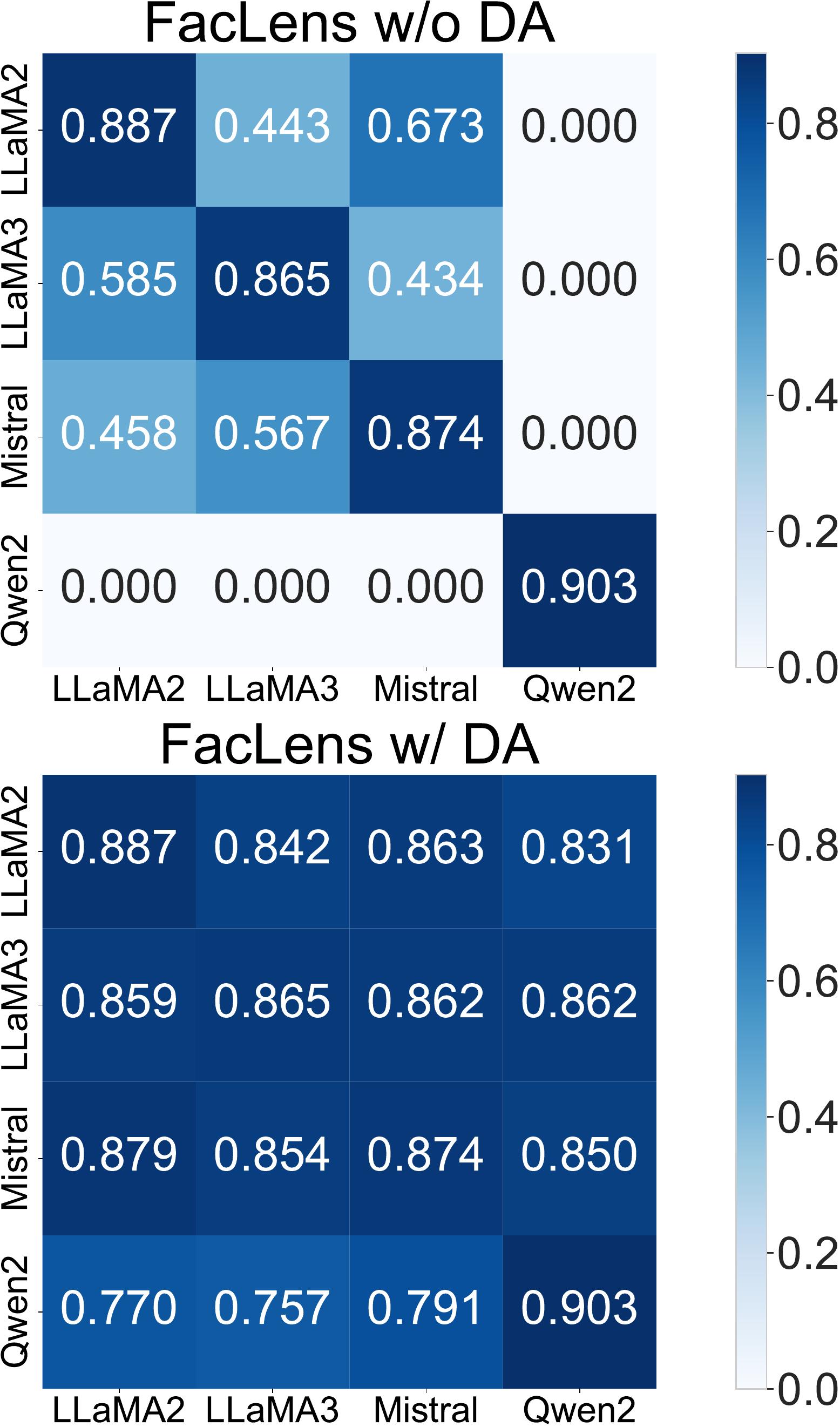}}
    \hspace{0.1in}
    \subfigure[AUC Results on EQ]{\includegraphics[width=0.25\textwidth]{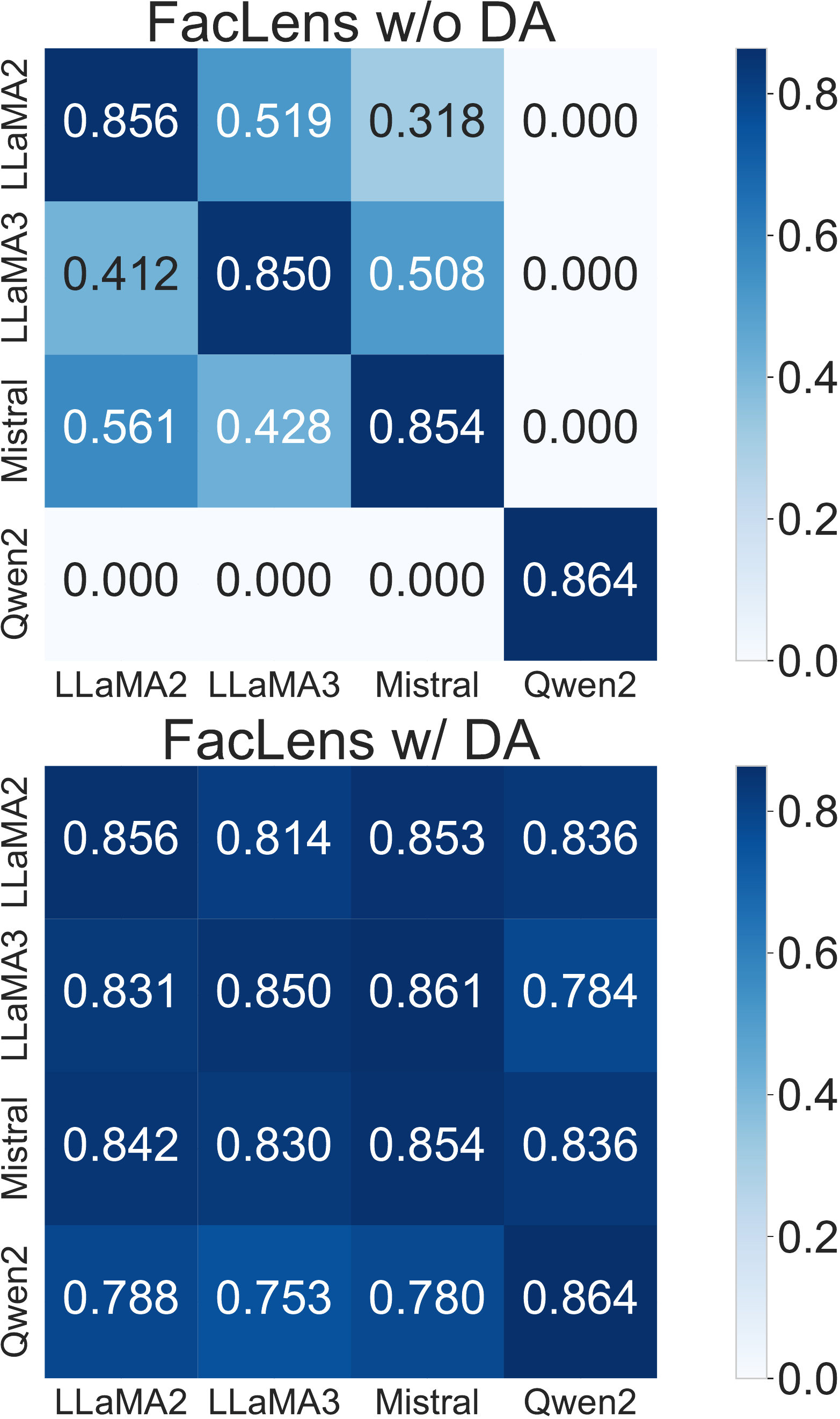}}
    \hspace{0.1in}
    \subfigure[AUC Results on NQ]{\includegraphics[width=0.25\textwidth]{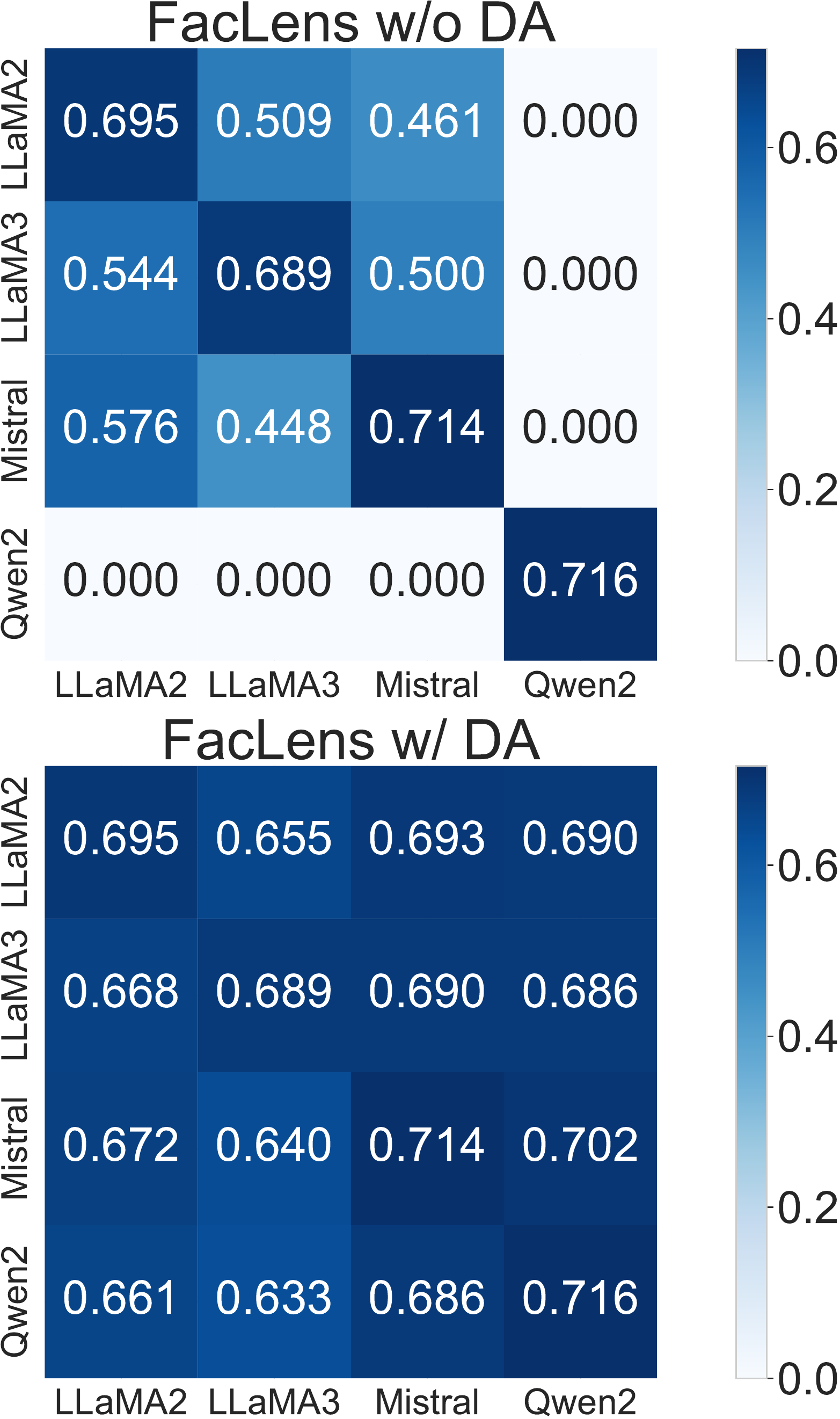}}
    \caption{Performance of cross-LLM \smodel w/o and w/ DA. In each heatmap, the element in the $i$-th row and $j$-th column represents the performance on the $j$-th target domain, with label information transferred from the $i$-th source domain.}
\label{fig:transfer_performance_linear_kernel}
\end{figure*}

\begin{figure*}[ht]
	\centering
		\includegraphics[width=1.9in]{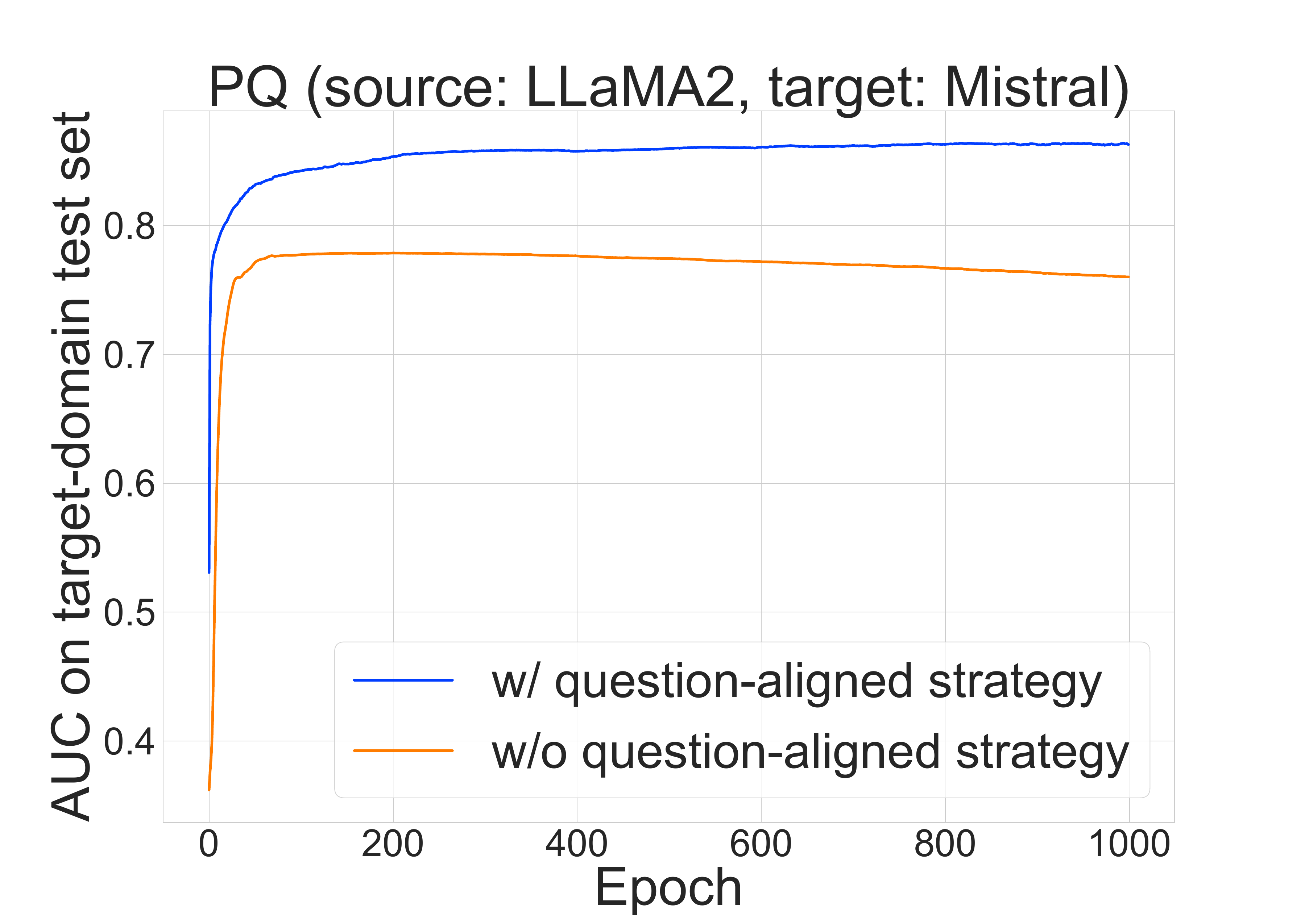}
          \includegraphics[width=1.9in]{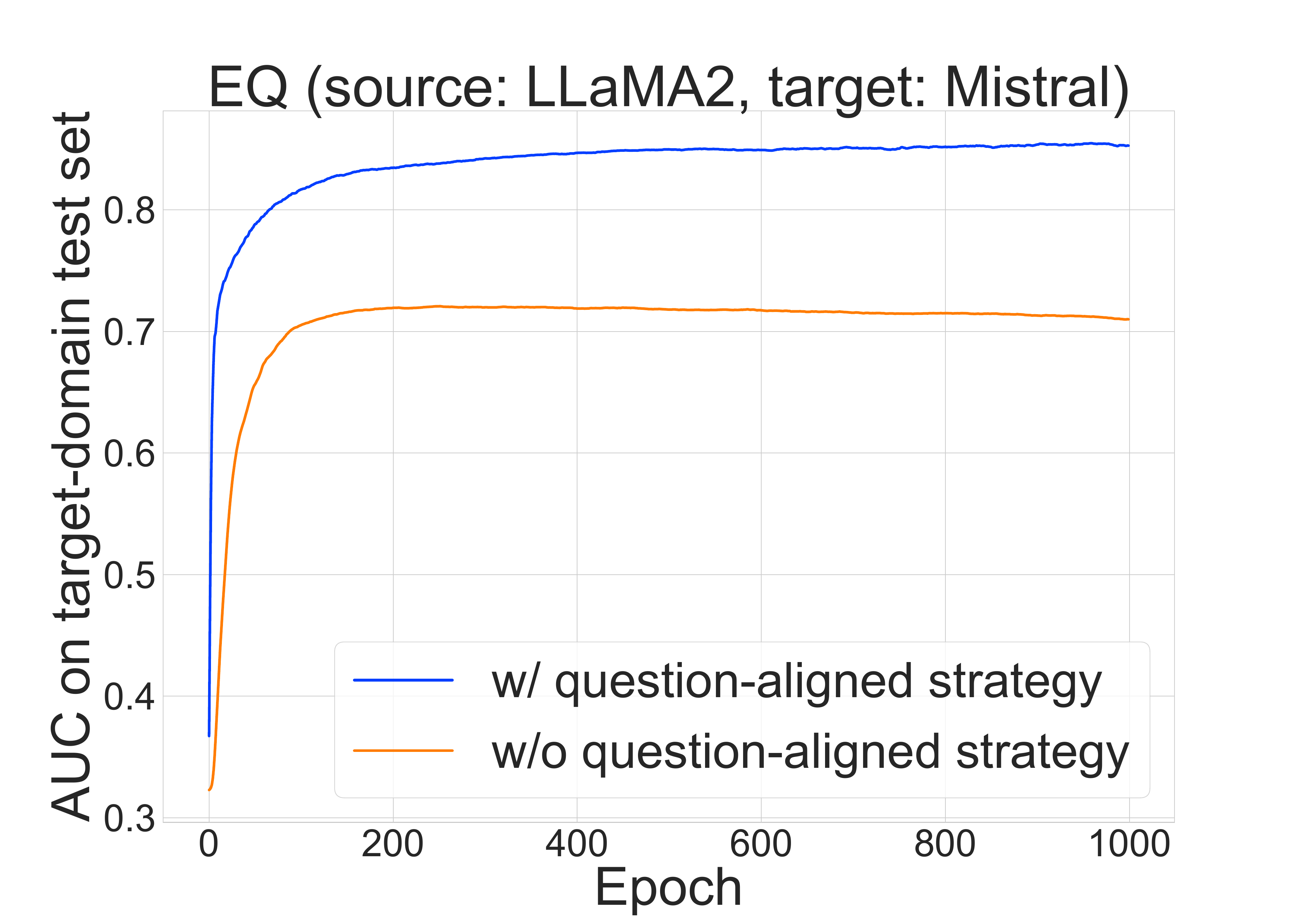}
		\includegraphics[width=1.9in]{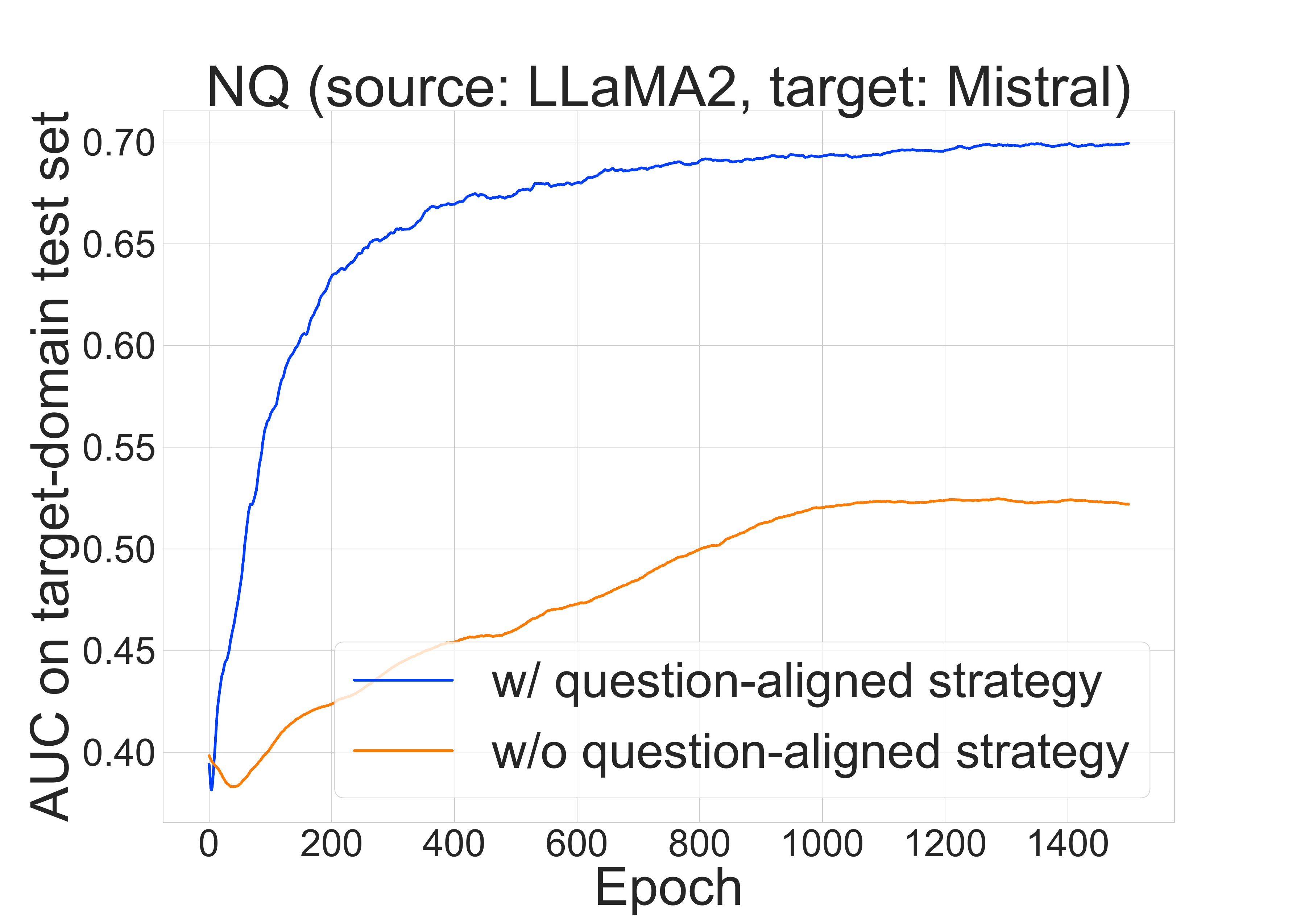}
	   \caption{Evaluation of the question-aligned mini-bath training strategy. Similar trends appear on other pairs of source and target domains.}
    \label{fig:question_aligned}
\end{figure*}

\textbf{\smodel stands out for its efficiency.}
Taking the LLaMA2-PQ NFP dataset as an example, which includes 2,272 questions for training, 1,136 questions for validation, and 7,952 questions for testing, Table~\ref{tab:efficacy} reports the average training time per epoch and the average prediction time per question for each method.
We can see that fine-tuning an LLM (i.e., Self-Evaluation and LoRA) incurs significant computational costs. 
Importantly, if the LLM has been fine-tuned for a specific task, its ability on other tasks can be compromised~\citep{GeneralizationOfFinetunedLLMs}.
As for prediction, \smodel runs much faster than Self-Familiarity because Self-Familiarity involves multiple conversations with the LLM.
As \smodel uses the middle-layer hidden question representations, it runs faster than Prompting, PPL, LoRA, and Self-Evaluation that involve more layers in the LLM.
SAT Probe extracts attention weights across all layers and attention heads, so \smodel runs faster during prediction.

\textbf{Unsupervised domain adaptation performs well for cross-LLM \model.}
Given an LLM, we train \smodel with the training data of the corresponding domain and directly test it on the test data of another domain.
The results in the upper part of Figure~\ref{fig:transfer_performance_linear_kernel} are unsatisfactory. 
After unsupervised DA (MMD loss with linear kernel), the cross-LLM \smodel can work much better in the target domain, as depicted in the the lower part of Figure~\ref{fig:transfer_performance_linear_kernel}.
Furthermore, we observe that \smodel shows better transferability between LLMs of similar scales. In future work, we will explore more effective methods to enhance \model's transferability between LLMs of very different scales.

\textbf{Question-aligned strategy is necessary to mini-batch training of cross-LLM \model.}
Figure~\ref{fig:question_aligned} shows that our question-aligned strategy for mini-batch training significantly enhances the performance of cross-LLM \model.
Particularly on the NQ dataset released by Google, which consists of questions from real users and covers more diverse questions, the estimation of $P_S\left(\mathbf{Z}\right)$ and $P_T\left(\mathbf{Z}\right)$  is more likely to be influenced by the sampling process in a mini-batch.
Hence, integrating the question-aligned strategy fosters the training process more on NQ.

\section{Conclusion}
In this paper, we find that the hidden representation of a fact-seeking question contains valuable information for identifying potential non-factual responses (i.e., NFP). We also discover that similar NFP patterns emerge in hidden question representations sourced from different LLMs. These findings support our lightweight and transferable NFP model, \model, which enables more efficient development and application. We hope this work can inspire future research on LLMs' factuality.

\clearpage
\section*{Limitations}
This paper assumes access to the parameters of LLMs, which limits the application of \smodel to API-based black-box LLMs. With the growing availability of advanced open-source LLMs, research on white-box methods is becoming increasingly important, making our work meaningful. 
However, we also recognize the necessity of designing effective and efficient NFP models for black-box LLMs in our future work.

\bibliography{custom}

\appendix
\begin{table*}[t]
   \small
   \centering
   \renewcommand\arraystretch{1}
    \newcolumntype{?}{!{\vrule width 1pt}}
    \begin{tabular}{l?cc?cc?cc}
        \toprule
        &\multicolumn{2}{c?}{\textbf{PQ}}&\multicolumn{2}{c?}{\textbf{EQ}}&\multicolumn{2}{c}{\textbf{NQ}} \\
        \cmidrule{2-7}
        & \textbf{\makecell[c]{Pos}}& \textbf{\makecell[c]{Neg (factual)}} & \textbf{\makecell[c]{Pos }}  & \textbf{\makecell[c]{Neg (factual)}}& \textbf{\makecell[c]{Pos }} & \textbf{\makecell[c]{Neg (factual)}} \\
        \midrule
        LLaMA2-7B-Chat & 74.9 & 25.1 & 70.3 & 29.7 & 57.2 & 42.8\\
        LLaMA3-8B-Instruct & 65.5 & 34.5 & 61.6 & 38.4 & 48.2 & 51.8\\
        Mistral-7B-Instruct-v0.2 & 73.0 & 27.0 & 68.2 & 31.8 & 55.5 & 44.5\\
        Qwen2-1.5B-Instruct & 86.2 & 13.8 & 80.1 & 19.9 & 75.9 & 24.1 \\
        \bottomrule
    \end{tabular}
    \caption{Positive and negative sample ratios in different NFP datasets (\%). A NFP dataset is built based on an LLM and a QA dataset.
    A positive (non-factual) sample indicates the LLM $m$ cannot provide the queried facts in response to the fact-seeking question $q$, whereas a negative (factual) sample indicates the LLM $m$ can provide the queried facts in response to the fact-seeking question $q$.}
    \label{tab:response accuracy}
\end{table*}

\begin{figure*}[h]
	\centering
		\includegraphics[width=6in]{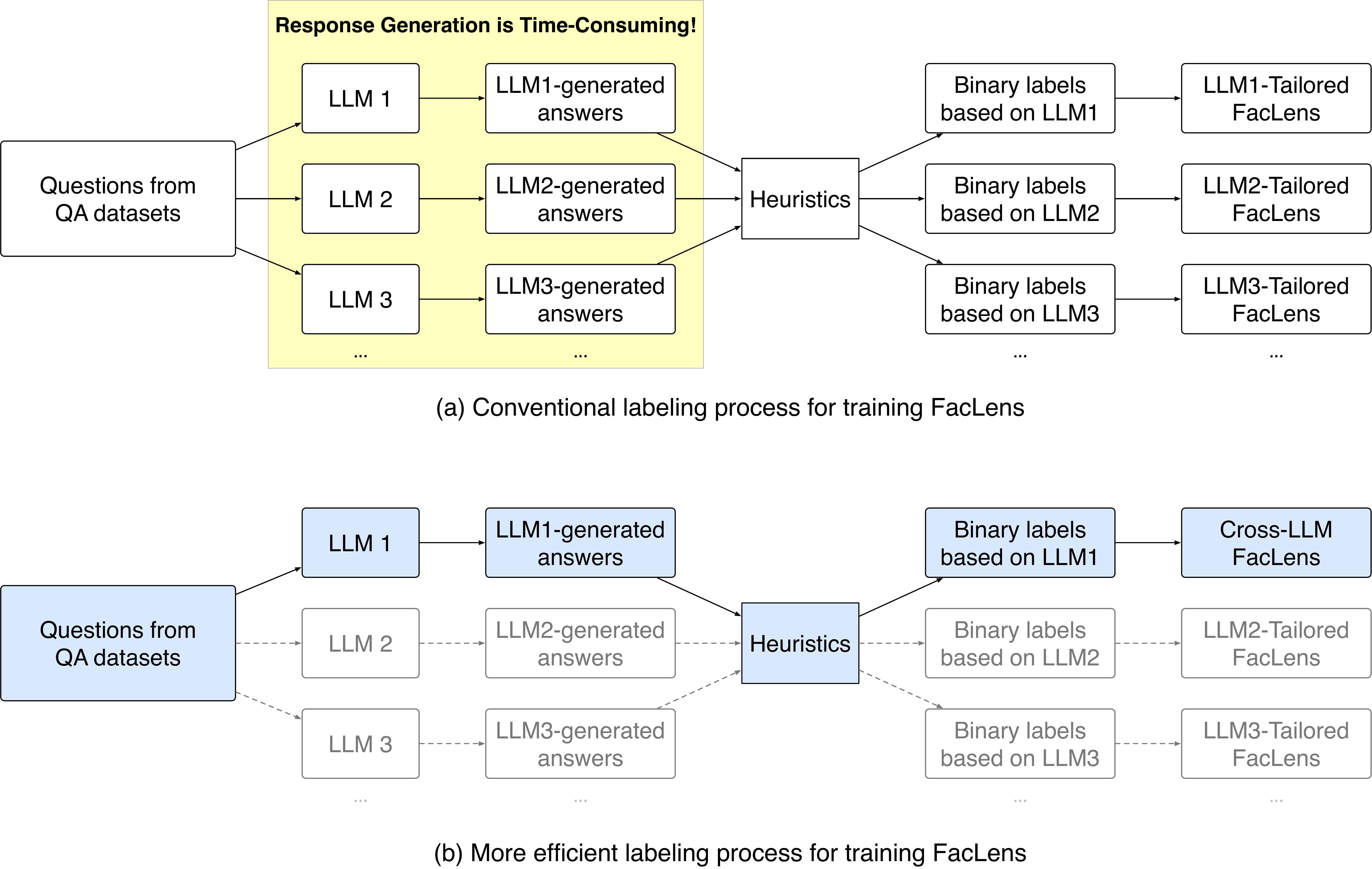}
  \caption{Illustration of different labeling processes for training \model.}
  \label{fig:labeling methods}
\end{figure*}

\section{Statistics of QA Datasets}\label{appendix:dataset}
During the NFP dataset construction, to reduce the false negative samples, we exclude multiple-choice questions because their LLM-generated responses are likely to mention both correct and incorrect answers.
Moreover, we exclude questions where the golden answers are three characters or fewer, as such short strings are likely to appear as substrings within unrelated words.
PQ initially contains 14,267 questions. After eliminating duplicates and removing the above special questions, 11,360 unique questions remained.
EQ contains 100K questions. We randomly sample 7,200 questions from EQ, ensuring uniform coverage across all question topics. After eliminating duplicates and removing special questions, we retain 7,159 questions from EQ.
The full NQ dataset is huge (about 42GB), so we download a simplified development set of NQ.
In this paper, we focus on the case of short answers, so we select questions whose answers are comprised of 30 characters or less. Hence, we include 1,244 questions sourced from NQ.
Table~\ref{tab:response accuracy} shows the ratios of positive and negative samples in each NFP dataset, where a pair of QA dataset and LLM corresponds to an NFP dataset.

\section{More Efficient Process of Training Data Construction}\label{appendix: Comparison Between Different Labeling Methods}
In Section~\ref{sec: pilot study transferability of faclens}, we analyzed why the transferability of \smodel can reduce overall development costs for multiple LLMs by lowering the costs of obtaining labels for \smodel training. Figure~\ref{fig:labeling methods} provides the illustration, where the gray dashed lines indicate that the corresponding steps are omitted.

\section{Experimental Settings}

\subsection{Perplexity (PPL) on a  Question}\label{appendix:form_of_PPL} 
We regard the PPL of a fact-seeking question as a baseline. In specific, we predict $y=1$ if the PPL value exceeds a certain threshold.
We extend the calculation of PPL to be conducted in each layer to obtain multiple PPL values for a text and determine the layer based on the NFP performance on labeled data.
Formally, PPL on a question calculated in the $\ell$-th layer is formulated as,
\begin{equation}
        \text{PPL} 
         = \exp\left(\frac{1}{|q|}\sum_{v_k \in q} 
        -\log\left(p_{\ell}\left(v_k | v_{<k}\right)\right) \right)
    \label{eq:latentperplexity}
\end{equation}

\begin{equation}
        p_{\ell}\left(v_k | v_{<k}\right) 
        = \text{Softmax}\left(m_{\leq \ell}\left(v_{<k}\right)W_{U}\right)_{v_k}
\end{equation}

where $q$ is a fact-seeking question, $v_k$ is the $k$-th token in $q$, $v_{<k}$ represents the set of tokens preceding the $k$-th token, and $W_{U}$ is the pre-trained unembedding matrix of the LLM $m$ that converts the hidden token representations into distributions over the vocabulary.

\subsection{Hyper-Parameter Settings}\label{appendix:baseline_experimental_setting}
Our experiments are conducted based on 4 * 80G NVIDIA Tesla A800 GPUs. We implement the encoder $g_{enc}$ of \smodel by a 3-layer MLP, setting the dimension of each MLP layer to 256. 
We use the Adam optimizer with weight decay 1e-4.
The hyper-parameters determined on the validation set include: the training epochs (set the maximum epochs to 100), and the learning rate $\in$ \{1e-3, 1e-4\} for single-LLM \model. Considering that the training questions from NQ are relatively small, we set the learning rate of \smodel to 1e-4 on NFP datasets derived from NQ.
The default learning rate of cross-LLM \smodel is set to 1e-5.
Due to the memory limitation, we minimize the MMD loss via mini-batch training with a batch size of 64.

In terms of baselines, we adopt hyper-parameter settings recommended by their authors.
Since we extend PPL to be calculated in each hidden layer, we determine the specific layer according to PPL's performance on the labeled data.
We introduce the Prompting-based method, which encourages an LLM to answer whether it knows the factual responses via prompt ``Question: \{question\}\textbackslash Can you provide a factual response to the above question? If you can, please reply yes or Yes. If you can not, please reply no or No.\textbackslash nAnswer: \{label\}\textbackslash n''.
The probabilities of predicting tokens ``yes'', ``Yes'', ``no'' and ``No'' are normalized for prediction.
For the Self-Evaluation (Fully FT), we train the model on 4*80G A800 GPUs, with a learning rate of 1e-6, batch size of 32, and epochs of 12, and we also determine the training epochs based on the performance on the validation set.
Self-Evaluation (Fully FT) needs to fully fine-tune an LLM. Therefore, to mitigate overfitting, the learning rate scheduler employs a cosine decay strategy with 5\% of the training steps dedicated to linear warm-up. Additionally, the final learning rate is set to one-tenth of its initial value. For LoRA, we integrate adapters on all ``q\_proj'', ``k\_proj'', ``v\_proj'', and ``o\_proj'' layers, while maintaining the original weights of the language model unchanged. The configuration is as follows: we specify a rank of 128 and an alpha of 256, with a learning rate of 1e-4, a batch size of 32, and the training is conducted over 32 epochs. We employ the same learning rate scheduler as used in Fully SFT. Because LoRA is a parameter-efficient fine-tuning technique, the training process requires only a single 80G A800 GPU.
Note that we chose hyperparameters (r=128 and alpha=256), which arr larger than those used in the original LoRA paper, to introduce more trainable parameters, thereby enhancing the modeling capacity of the LoRA adapters. Although these values are larger than those used in the original LoRA paper, the additional trainable parameters remain significantly smaller than those of the original LLM. For instance, with LLaMA2, LoRA adapters with r=128 and alpha=256 introduce only 1.95\% trainable parameters.

\begin{figure*}[h]
	\centering
          \includegraphics[width=2.3in]{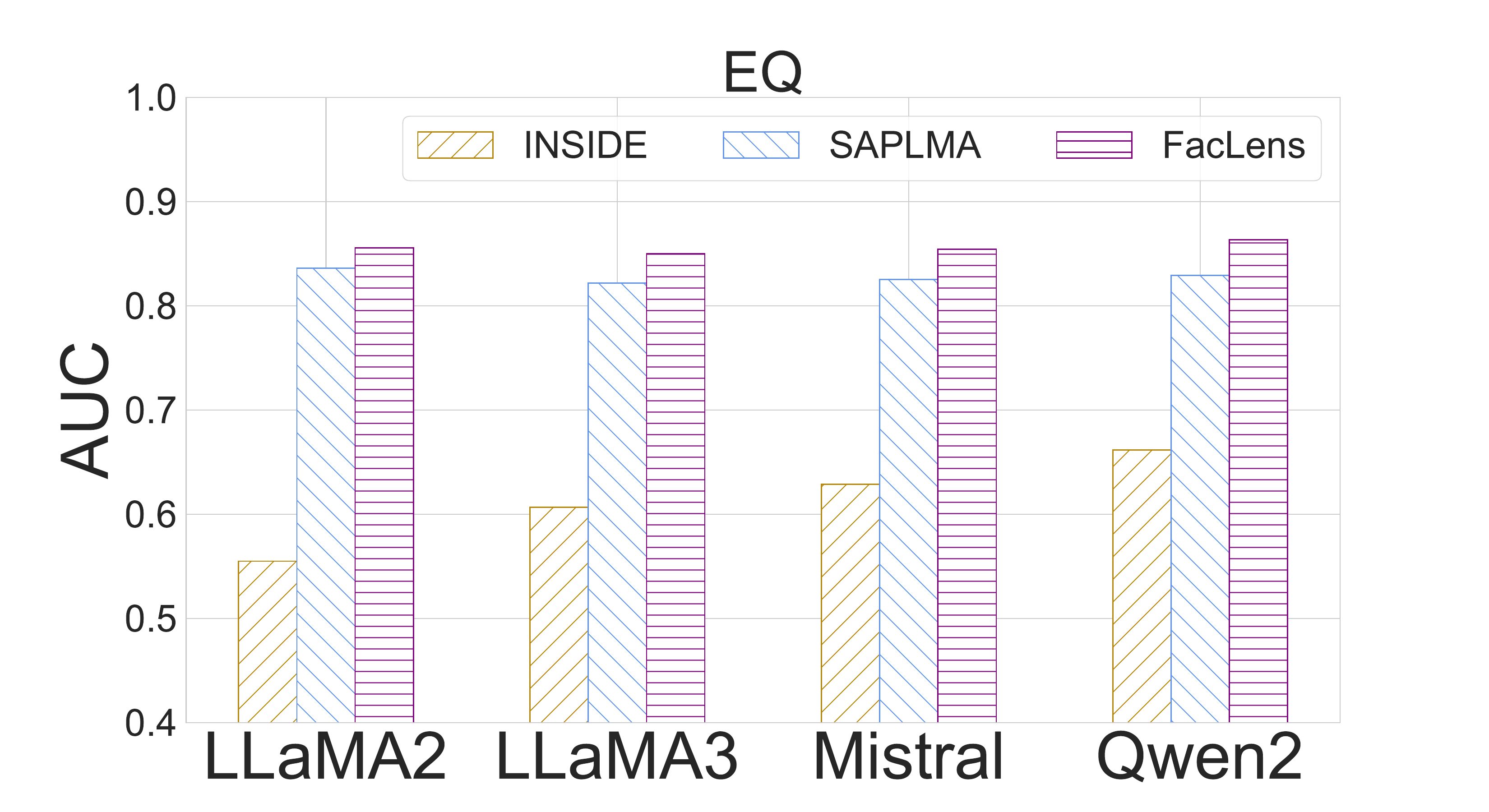} 
		\includegraphics[width=2.3in]{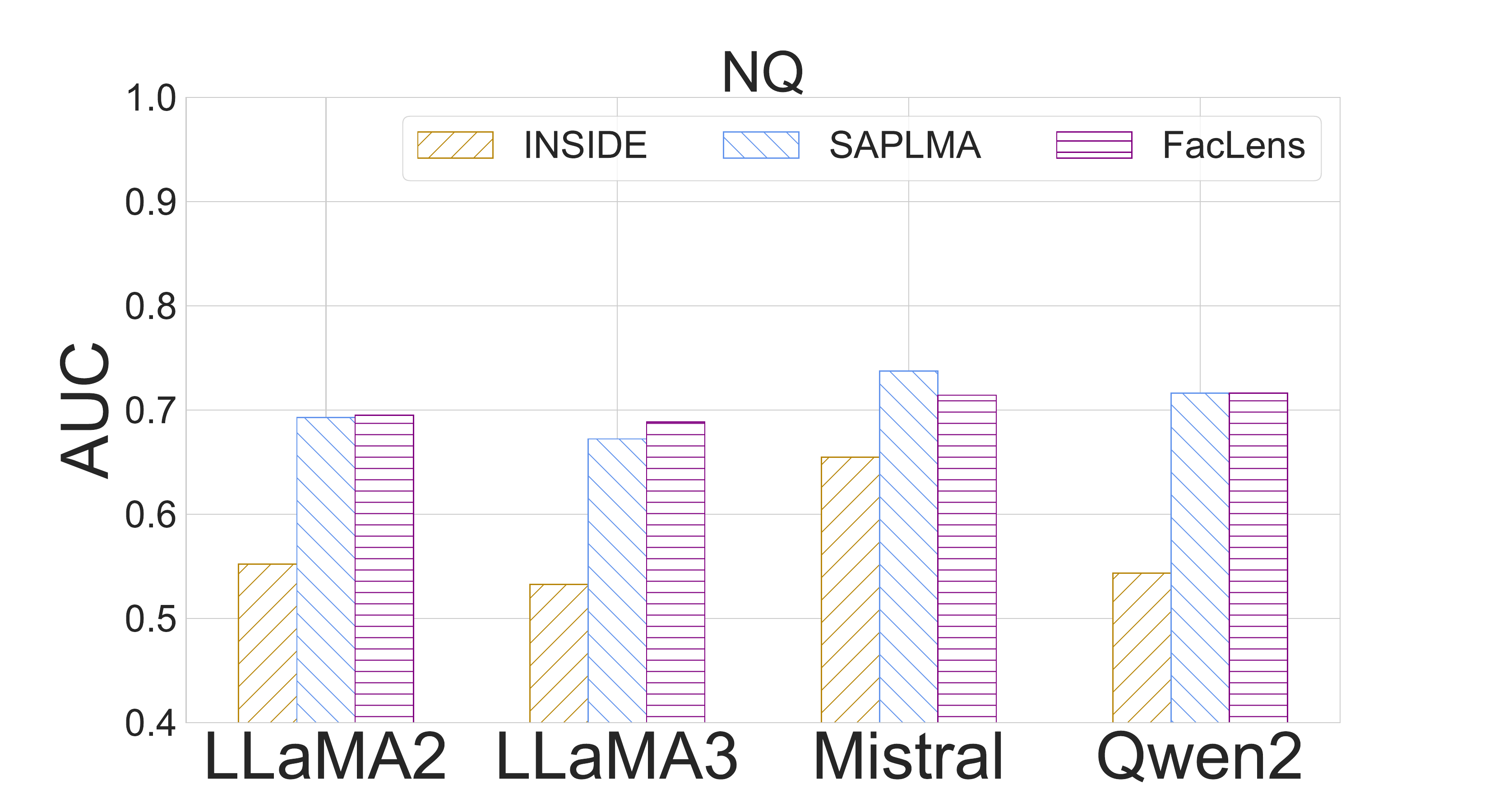}
	   \caption{Performance comparison of \smodel (ante-hoc NFP) with INSIDE and SAPLMA (post-hoc NFD) on EQ and NQ.}
    \label{fig:compare_with_post_hoc_NFD_EQ_NQ}
\end{figure*}

\begin{figure*}[h]
	\centering
          \includegraphics[width=2.3in]{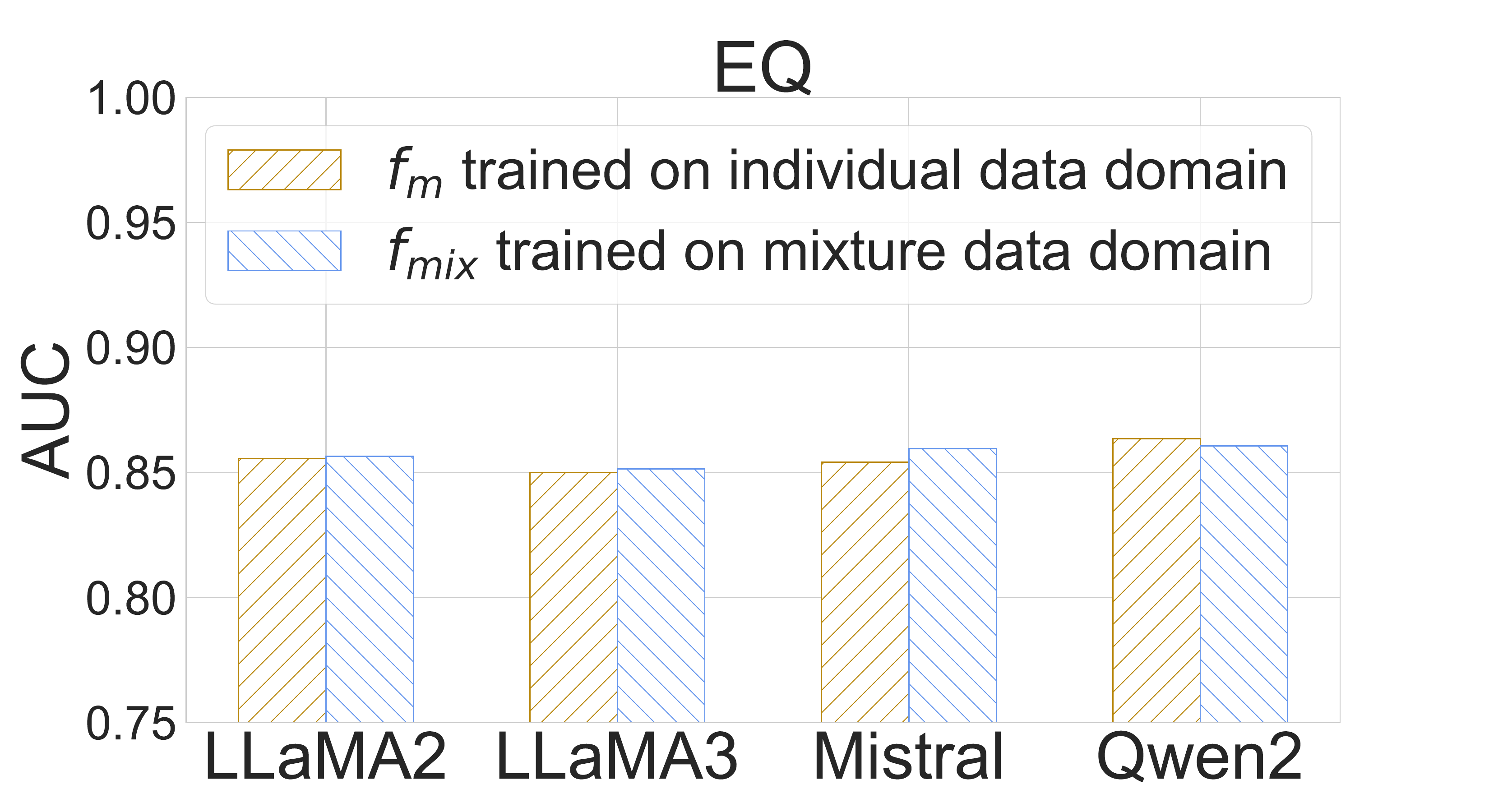}
		\includegraphics[width=2.3in]{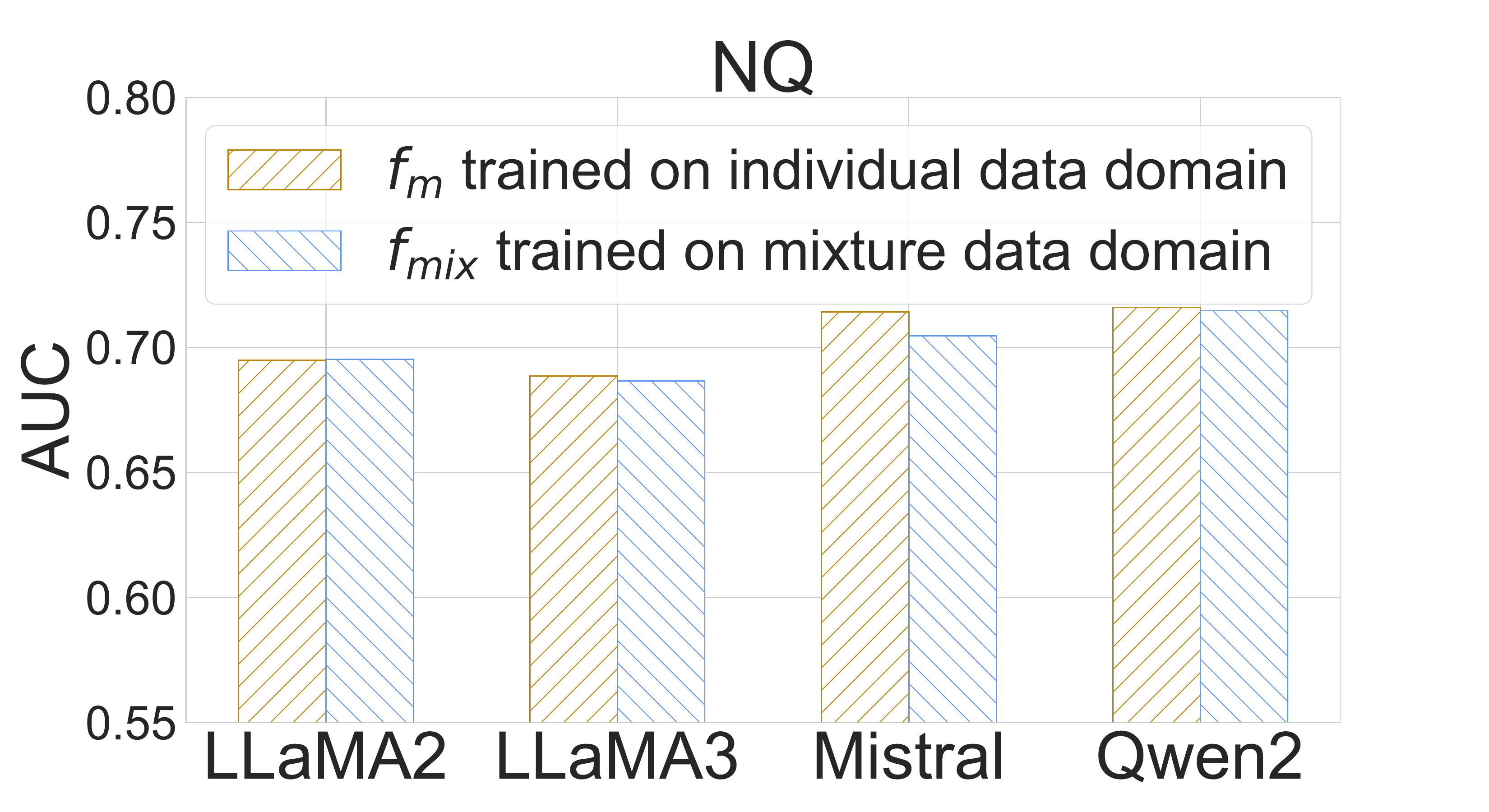}
	   \caption{Performance comparison between $f_{m}$ and $f_{mix}$ on EQ and NQ. Similar performance suggests no significant concept shift across different domains.}
    \label{fig:pilot_study_for_DA_EQ_NQ}
\end{figure*}

\begin{figure}[h]
	\centering
            \includegraphics[width=2.1in]{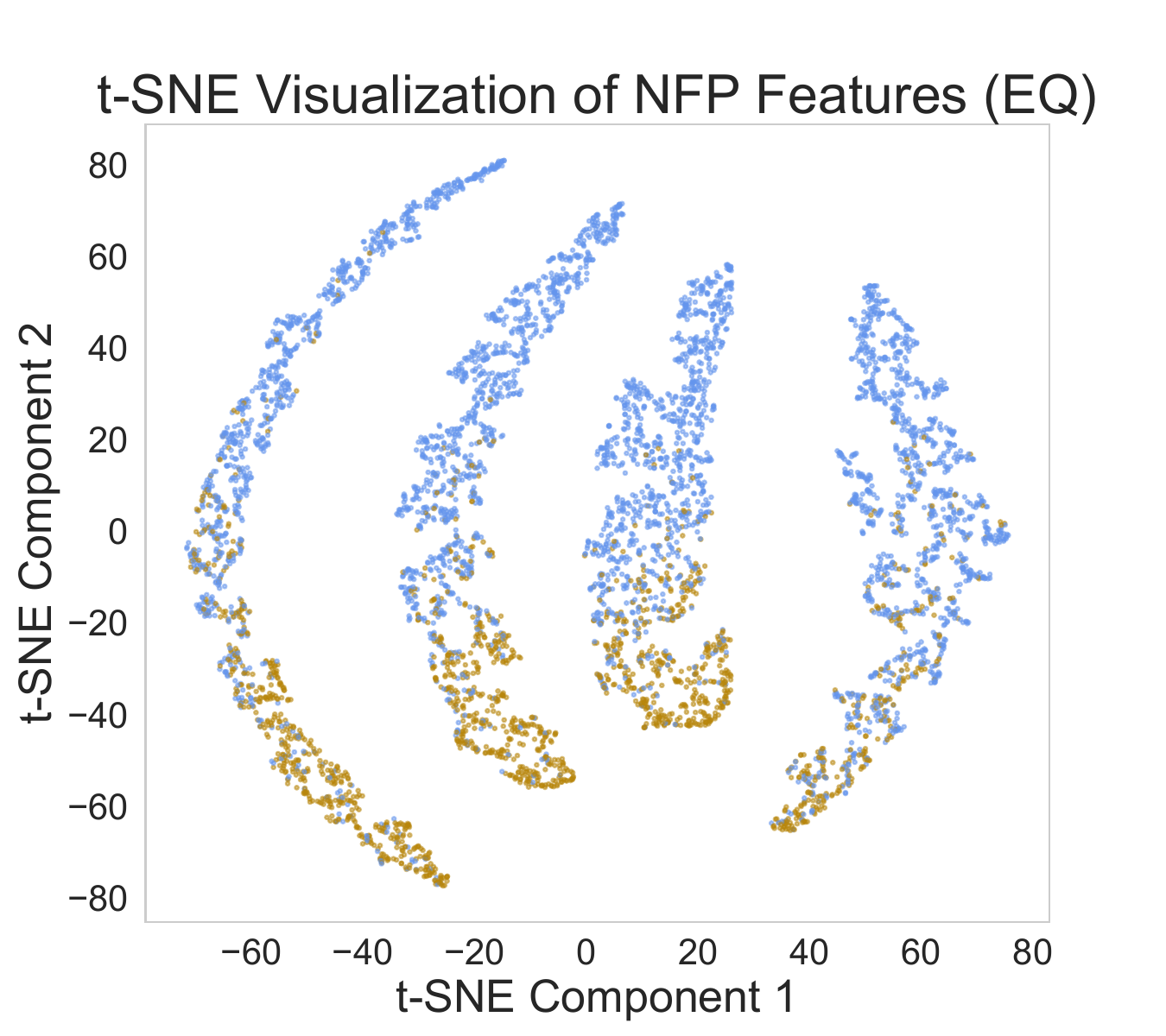}
          \includegraphics[width=2.1in]{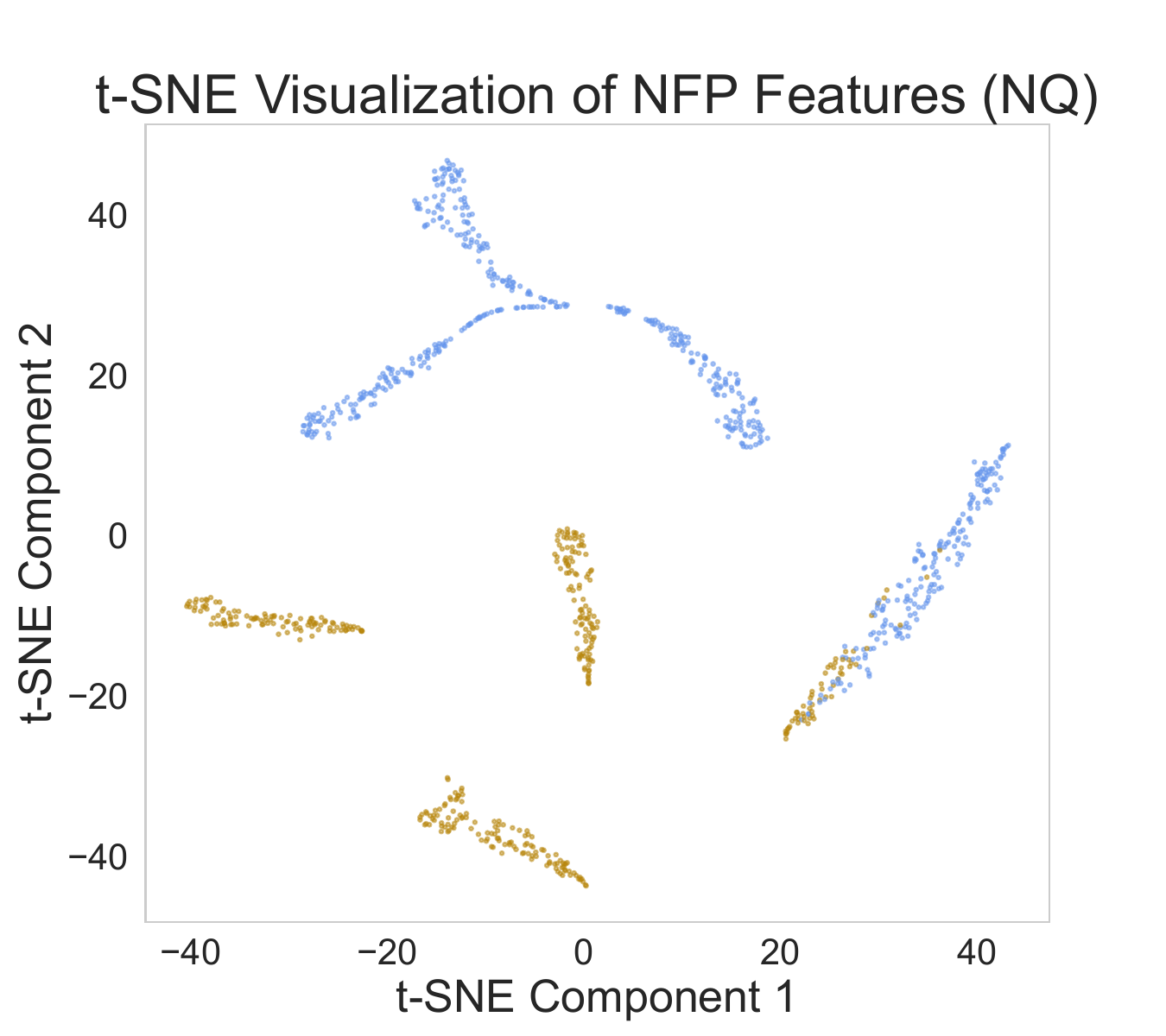}
	   \caption{Visualizations of NFP features.}
       \label{fig:pilot_study_for_DA_visualization_EQ_NQ}
\end{figure}

\subsection{Why Certain Baselines Are Unsuitable for the QA Datasets}\label{appendix: baseline not suitable for some datastes}
In Table~\ref{tab:exp_overall_evaluation}, Entity-Popularity and SAT Probe are not suitable for certain datasets.
Here, we explain the reasons.
Entity-Popularity uses Wikipedia page views to approximate the entity's popularity. However, EQ and NQ datasets do not provide relevant Wikipedia page views, and not every subject entity in the two datasets can be uniquely matched to a Wikidata entity. As a result, Entity-Popularity is unsuitable for EQ and NQ.
For the baseline SAT Probe, each question is assumed to contain constraint tokens, and the model extracts LLMs' attention to the constraint tokens to probe factuality. The authors of SAT Probe have restricted the formats of questions to directly identify the constraint tokens. However, extracting constraint tokens from free-form questions can be challenging. For PQ and EQ, which are template-based, obtaining constraint tokens is relatively straightforward. However, SAT Probe is not suitable for NQ, as questions in NQ come from real users and exhibit diverse structures.

The core of the SAT Probe is using an LLM’s attention weights to constraint tokens within a question to reflect the LLM's factual accuracy.
In the original paper, the SAT probe is implemented by a linear layer, optimized by logistic regression. To compare the effectiveness of hidden representations and attention weights, we employ the same MLP structure and CE loss for both SAT Probe and \model.

\section{Supplementary Experiments}

\subsection{Ante-Hoc NFP vs. Post-Hoc NFD}\label{appendix: Ante-Hoc NFP vs. Post-Hoc NFD}
In Figure~\ref{fig:compare_with_post_hoc_NFD_EQ_NQ}, we supplement the performance comparison between hidden representation-based ante-hoc and post-hoc methods on EQ and NQ. The results further highlight that \smodel (ante-hoc NFP method) has the potential to outperform the post-hoc NFD methods.

\begin{table*}[h]
   \small
   \centering
   \renewcommand\arraystretch{1.1}
   \begin{threeparttable}
    \newcolumntype{?}{!{\vrule width 1pt}}
    \setlength{\tabcolsep}{1.8mm}
    \begin{tabular}{l?ccc?ccc?ccc?ccc}
        \toprule
        &\multicolumn{3}{c?}{\textbf{LLaMA2}}&\multicolumn{3}{c?}{\textbf{LLaMA3}}&\multicolumn{3}{c?}{\textbf{Mistral}} & \multicolumn{3}{c}{\textbf{Qwen2}}\\
        \cmidrule{2-13}
        & \textbf{PQ}& \textbf{EQ} & \textbf{NQ}  & \textbf{PQ}& \textbf{EQ} & \textbf{NQ} 
 & \textbf{PQ}& \textbf{EQ} & \textbf{NQ} & \textbf{PQ}& \textbf{EQ} & \textbf{NQ} \\
        \midrule        
        \smodel (avg, last layer) & 87.9 & 84.8 & 63.8 & 84.2 &  82.4 & 60.6 & 86.9 & 85.3 & 63.3 & 90.1 & 84.8 & 70.6\\
        \smodel (avg, 2$^{\text{nd}}$ to last layer) & 87.5 & 85.1 & 59.9 & 84.5 & 83.0 & 54.9 & 87.4 & 85.8 & 64.9 & 89.6 & 84.4 & 70.7\\
        \smodel (avg, middle layer) & 88.5 & 85.9 & 66.0 &  85.5 & 84.8 & 62.8 & 87.5 & 84.7 & 67.6 & 89.0 & 86.2 & 70.8\\
        \bottomrule
    \end{tabular}
    \end{threeparttable}
    \caption{Prediction performance of \smodel (avg) (AUC \%).}
    \label{tab:evaluation_faclens_avg}
\end{table*}

\begin{table*}[h]
    \small
   \centering
   \renewcommand\arraystretch{1}
   \begin{threeparttable}
    \newcolumntype{?}{!{\vrule width 1pt}}
    \setlength{\tabcolsep}{1.8mm}
    \begin{tabular}{l?c?c?c?c}
        \toprule
        &\textbf{LLaMA2}& \textbf{LLaMA3}&\textbf{Mistral}& \textbf{Qwen2}\\
        \midrule
         PPL & 55.2 & 55.2 & 54.4 & 53.5\\
         Prompting & 62.5 & 61.0 & 63.0 & 62.1\\
         Entity-Popularity& -- & -- & -- & --\\
         SAT Probe & -- & -- & -- & --\\
         Self-Familiarity & 55.3 & 56.8 & 54.5 & 53.7\\
         LoRA (Parameter-Efficient FT)& 72.9 & 68.0 & 66.9 & 70.8\\
         Self-Evaluation (Fully FT)& 75.0 & 69.3 & 71.1 & 72.1\\
        
         \midrule
         
        \smodel (last token, last layer) & 74.3 & \textbf{68.6} & 74.1 & 72.1\\
        \smodel (last token, 2$^{\text{nd}}$ to last layer) & 74.7 & 68.3 & 74.7 & \textbf{72.7}\\
        \smodel (last token, middle layer) & \textbf{75.5} & 66.9 & \textbf{74.9} & 71.4\\
        \bottomrule
    \end{tabular}
    \end{threeparttable}
    \caption{Prediction performance of different NFP methods on HotpotQA (AUC \%).}
    \label{tab:exp_overall_evaluation_hotpotqa}
\end{table*}

\subsection{Demonstrations of \model's Transferability}
\label{appendix: Demonstration of FacLens's Transferability}
In Section~\ref{sec: pilot study transferability of faclens}, we compare \smodel $f_{m}$, trained on an individual domain, with \smodel $f_{mix}$, trained on the mixture domain. The results show that $f_{mix}$ exhibits comparable performance to $f_{m}$ on the test set of the corresponding individual domain. We supplement the results based on questions from EQ and NQ in Figure~\ref{fig:pilot_study_for_DA_EQ_NQ}, further demonstrating no significant concept shifts between domains.

Besides, we supplement more visualizations of the NFP features in Figure~\ref{fig:pilot_study_for_DA_visualization_EQ_NQ}, further demonstrating that a unified classification boundary can be
applied to different LLMs in the NFP task.

\subsection{Performance of \smodel (avg)}\label{appendix:FacLens(avg)}
We use the averaged hidden representation of all tokens in a question as input to \model, denoted as \smodel (avg). Comparing the results in Table~\ref{tab:exp_overall_evaluation} and Table~\ref{tab:evaluation_faclens_avg}, we observe that \smodel (last token) performs more stably. Therefore, we recommend using the hidden representation of the last token in a question as the hidden question representation.

\begin{figure}[h]
	\centering
		\includegraphics[width=2in]{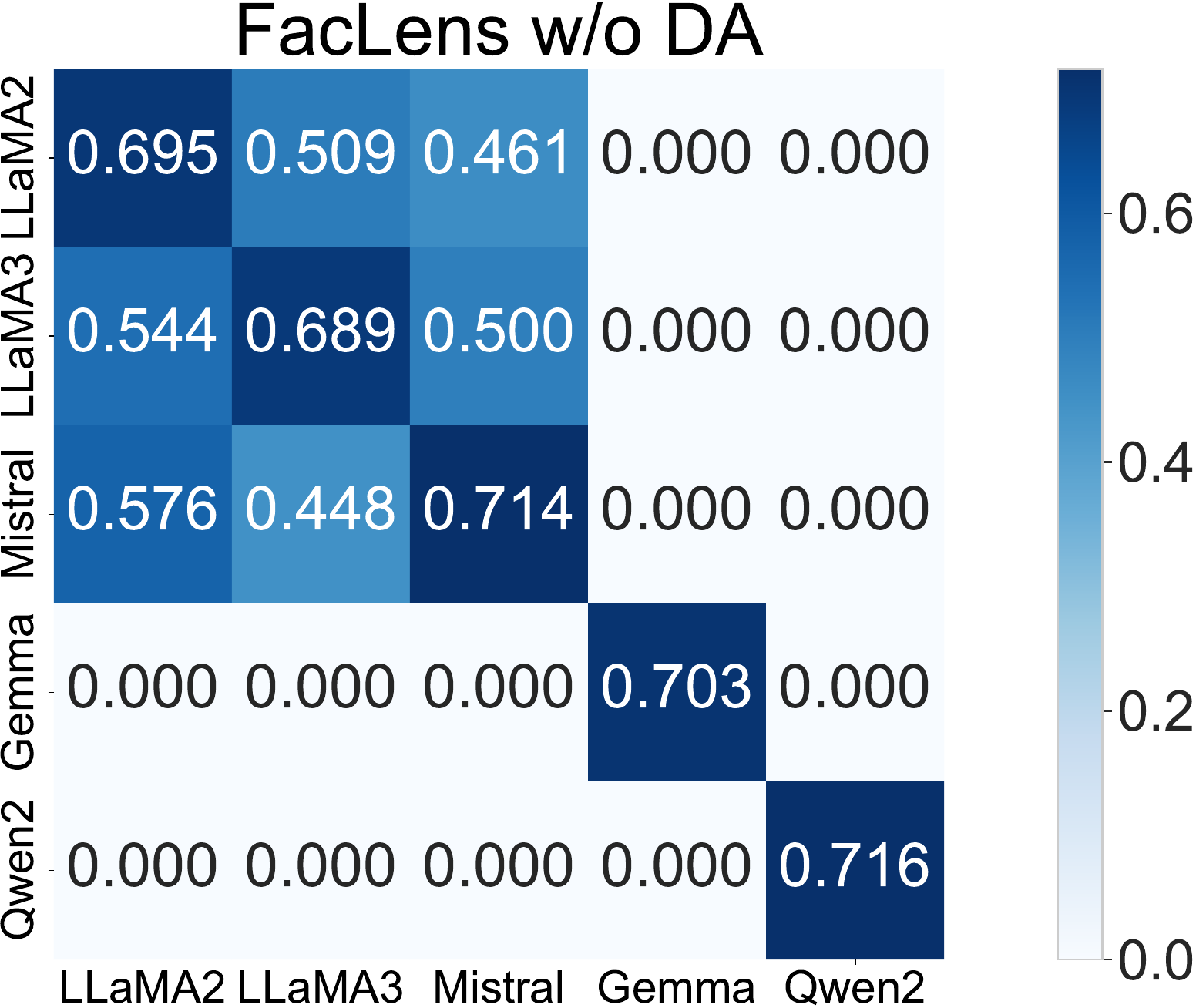}
		\includegraphics[width=2in]{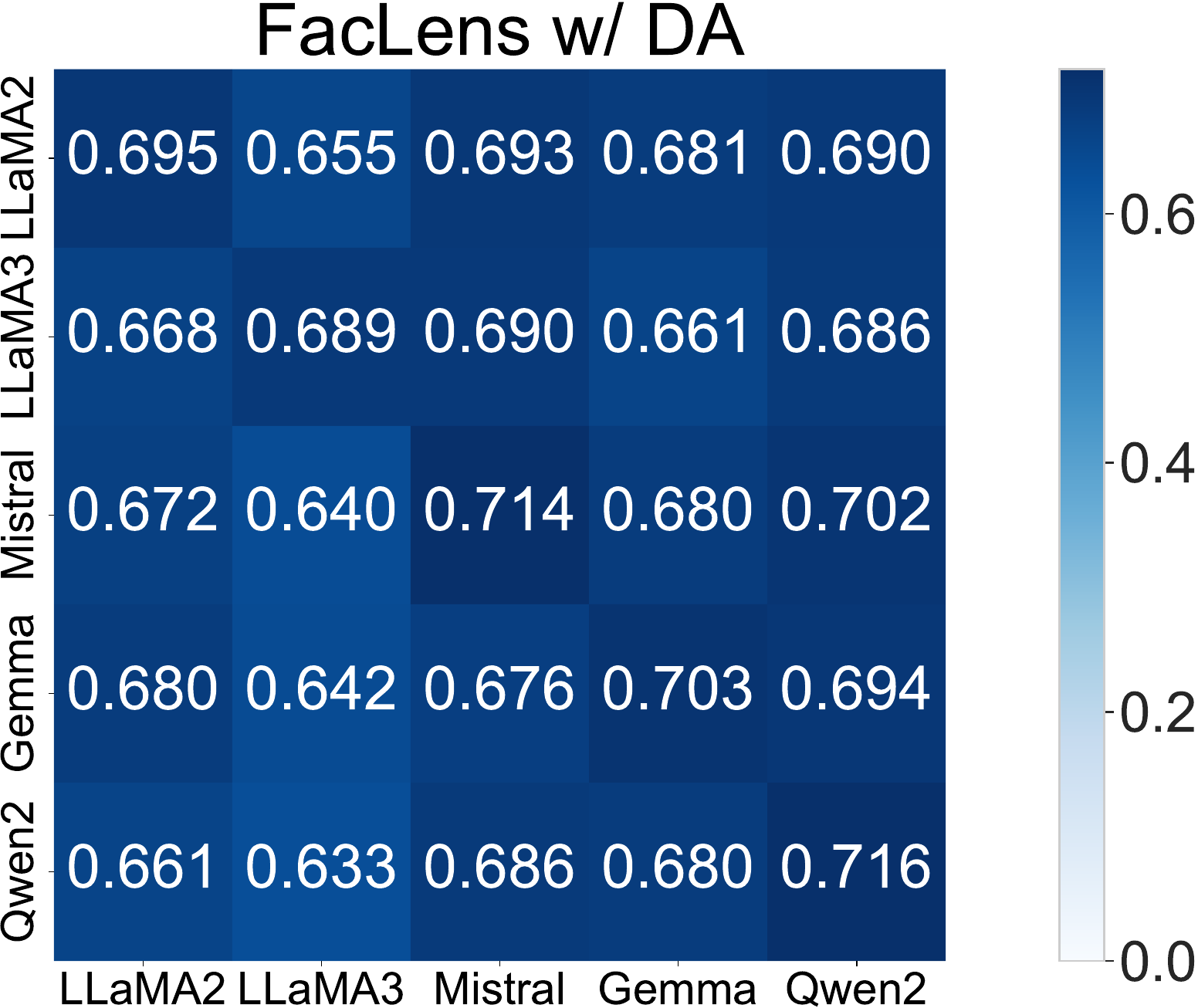}
	   \caption{Evaluation of cross-LLM \smodel on LLMs of different hidden dimensions. Questions are from NQ.}
    \label{fig:DA different hidden dimenssions}
\end{figure}

\subsection{Evaluation on HotpotQA}\label{appendix: Evaluation on the More Complex QA Dataset HotpotQA}

Table~\ref{tab:exp_overall_evaluation_hotpotqa} shows the prediction performance of different NFP methods on HotpotQA.
The observations are consistent with that in Table~\ref{tab:exp_overall_evaluation}, indicating \smodel can also handle the more complex fact-seeking questions.

\subsection{Cross-LLM \smodel for LLMs of Distinct Hidden Dimensions}\label{appendix: cross-LLM faclens different dimension}
Both Qwen2-1.5B-Instruct~\citep{qwen2} and Gemma-7B-it~\citep{Gemma} have different hidden dimensions compared to LLaMA2-7B-Chat, LLaMA3-8B-Instruct, and Mistral-7B-Instruct-v0.2.
The hidden dimension of Qwen2-1.5B-Instruct is 1536, and the hidden dimension of Gemma-7B-it is 3072, while the hidden dimension of LLaMA2-7B-Chat, LLaMA3-8B-Instruct, and Mistral-7B-Instruct-v0.2 is 4096.
A \smodel specially trained for the source-domain LLM cannot be directly used for a target-domain LLM whose hidden dimension is distinct from that of the source-domain LLM. Hence we introduce a linear layer to reshape the target-domain hidden question representations to match the dimension of the source domain’s hidden question representations and still adopt Eq.~\ref{eq:DA} to conduct domain adaptation.
In Figure~\ref{fig:DA different hidden dimenssions}, we can see that although two LLMs have different hidden dimensions, the cross-LLM \smodel can work well.

\begin{figure*}[h]
    \centering
		
        \includegraphics[width=1.85in]{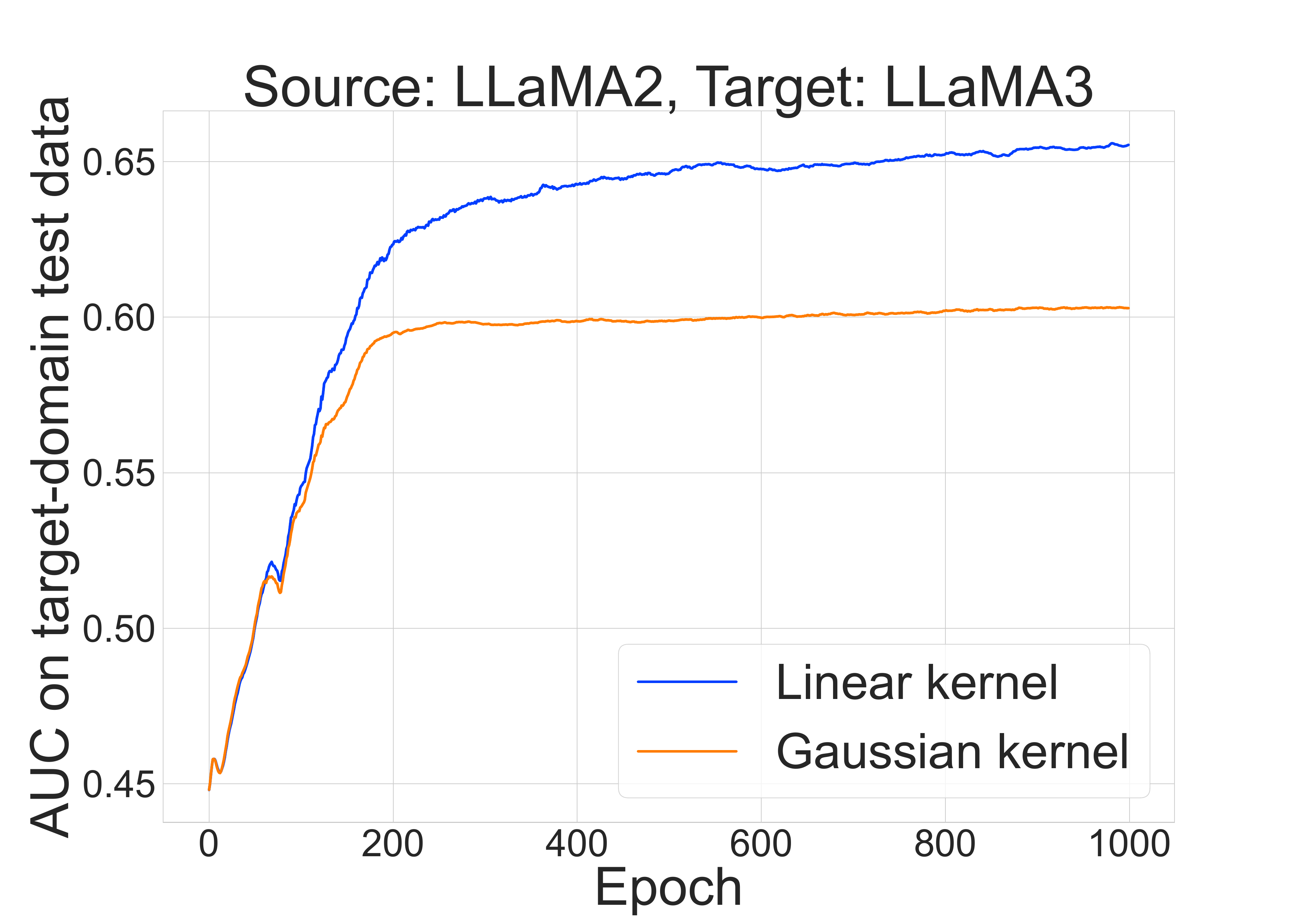}
          \includegraphics[width=1.85in]{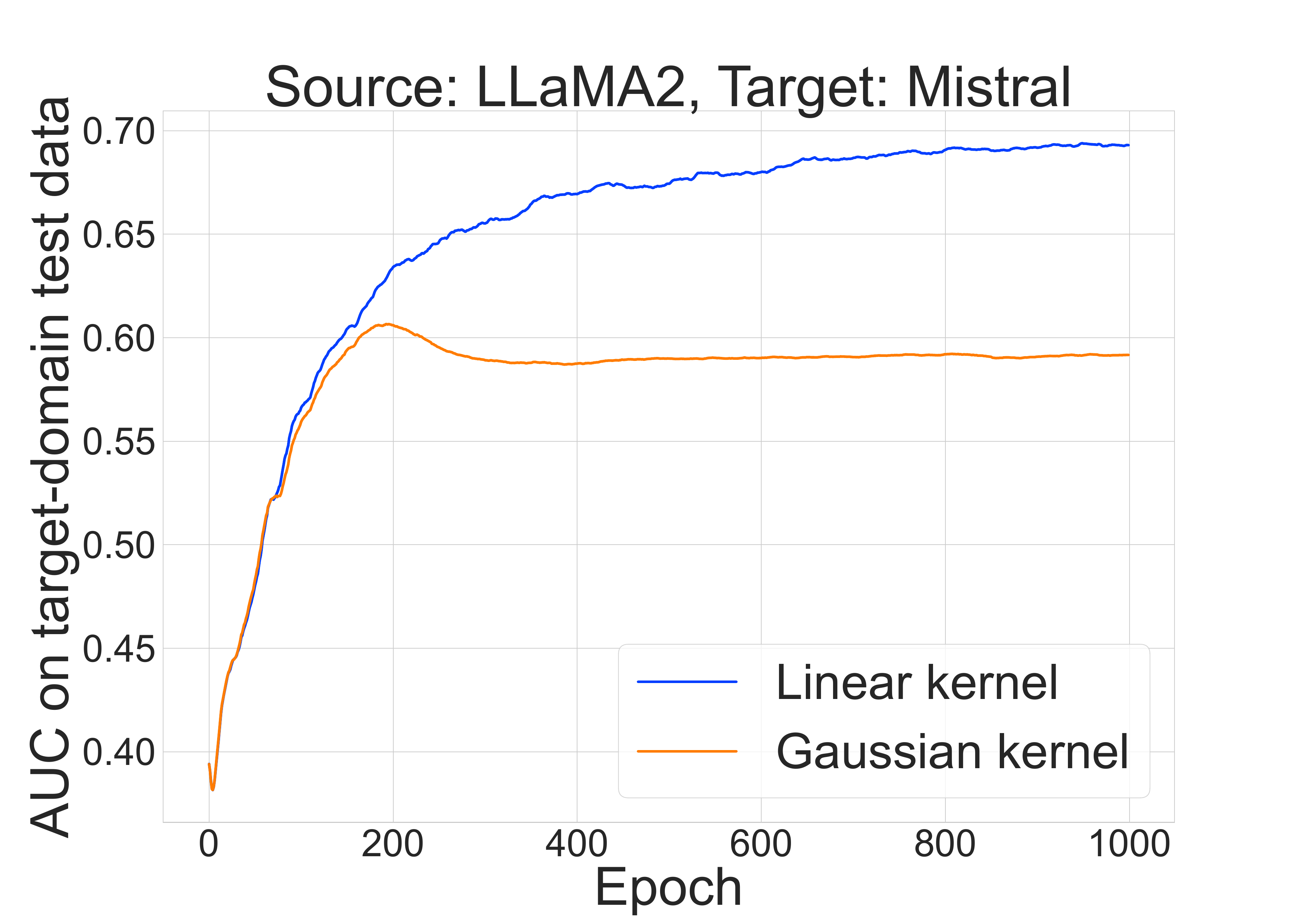}
          \includegraphics[width=1.85in]{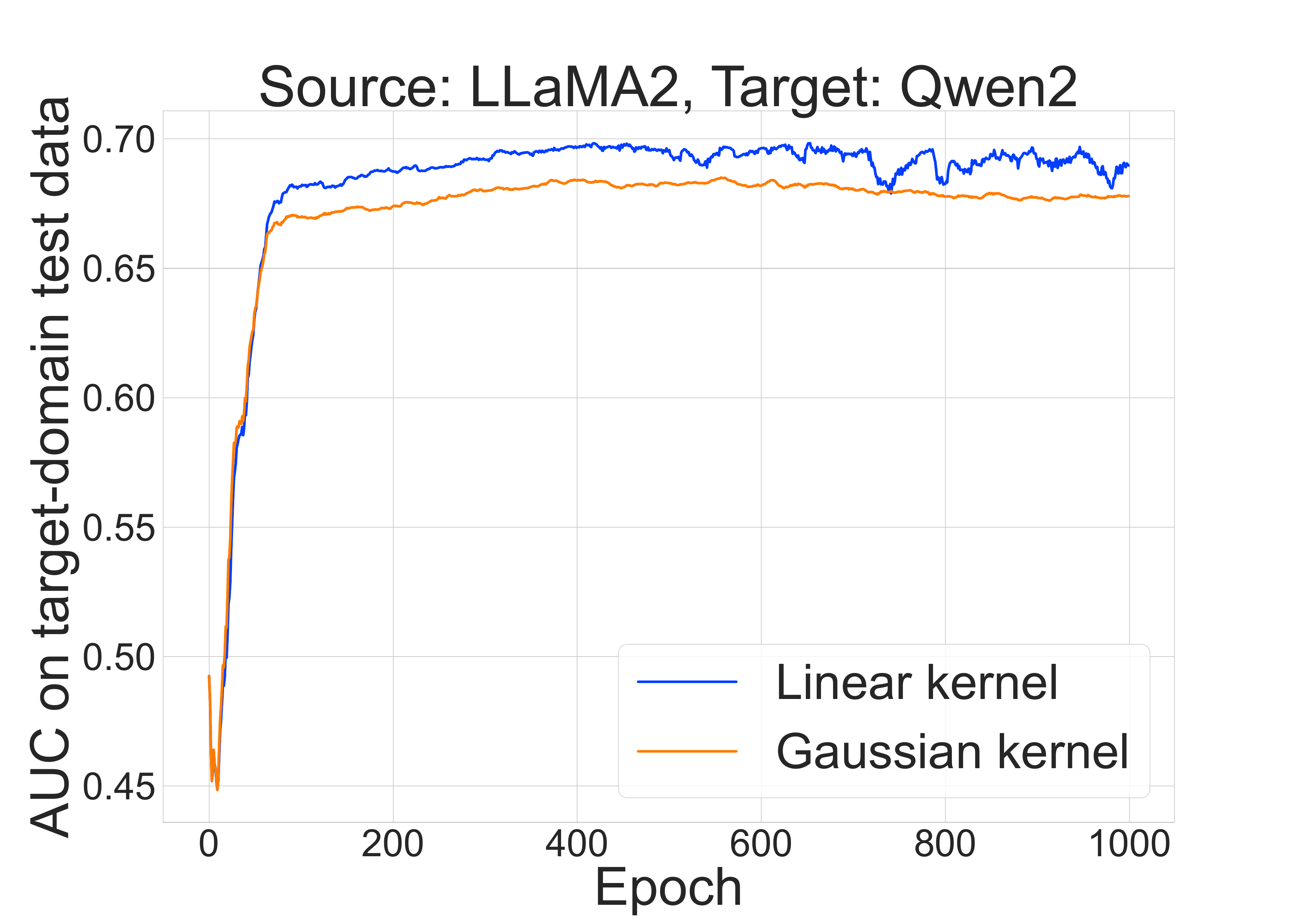}
            \includegraphics[width=1.85in]{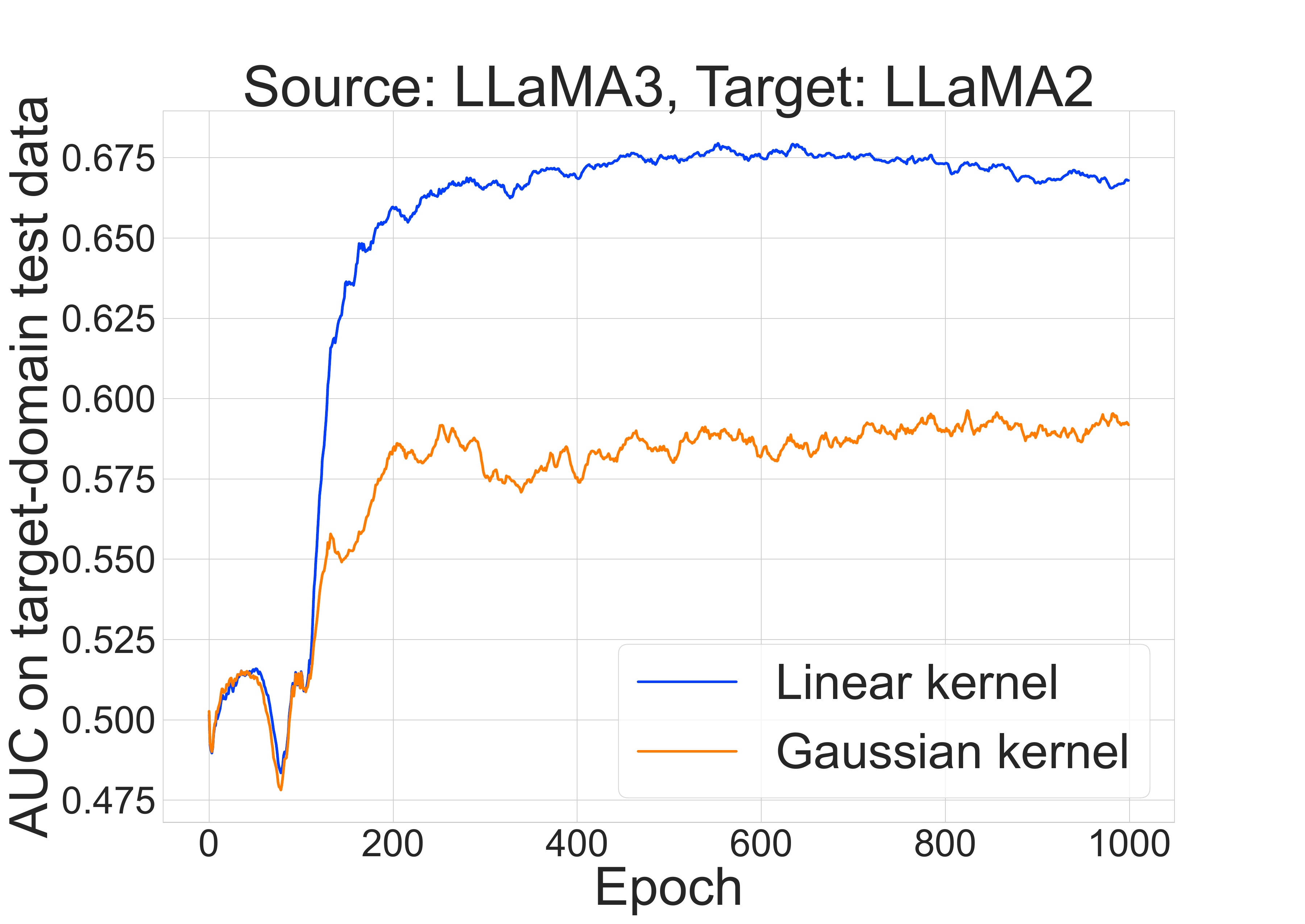}
        \includegraphics[width=1.85in]{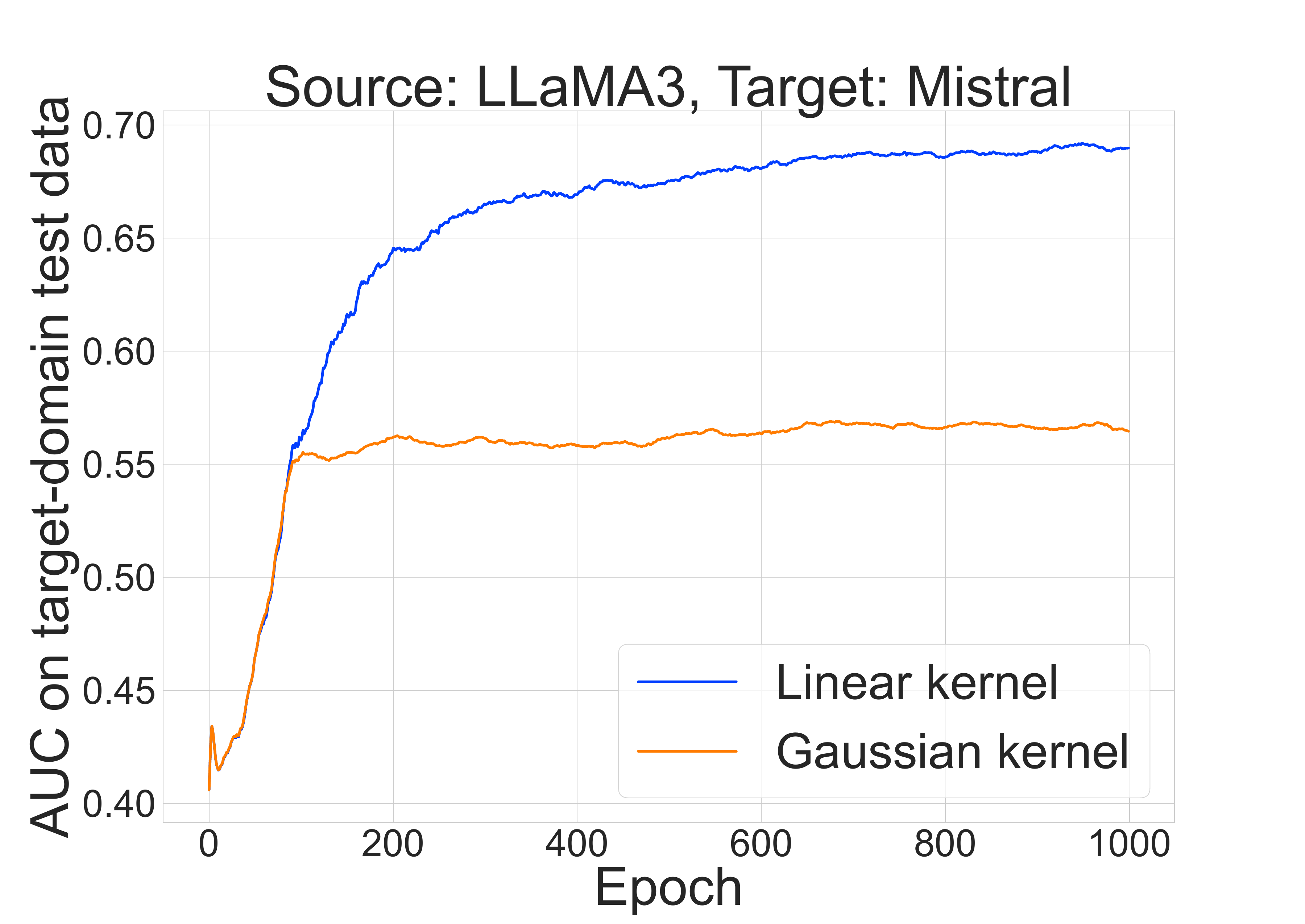}
        \includegraphics[width=1.85in]{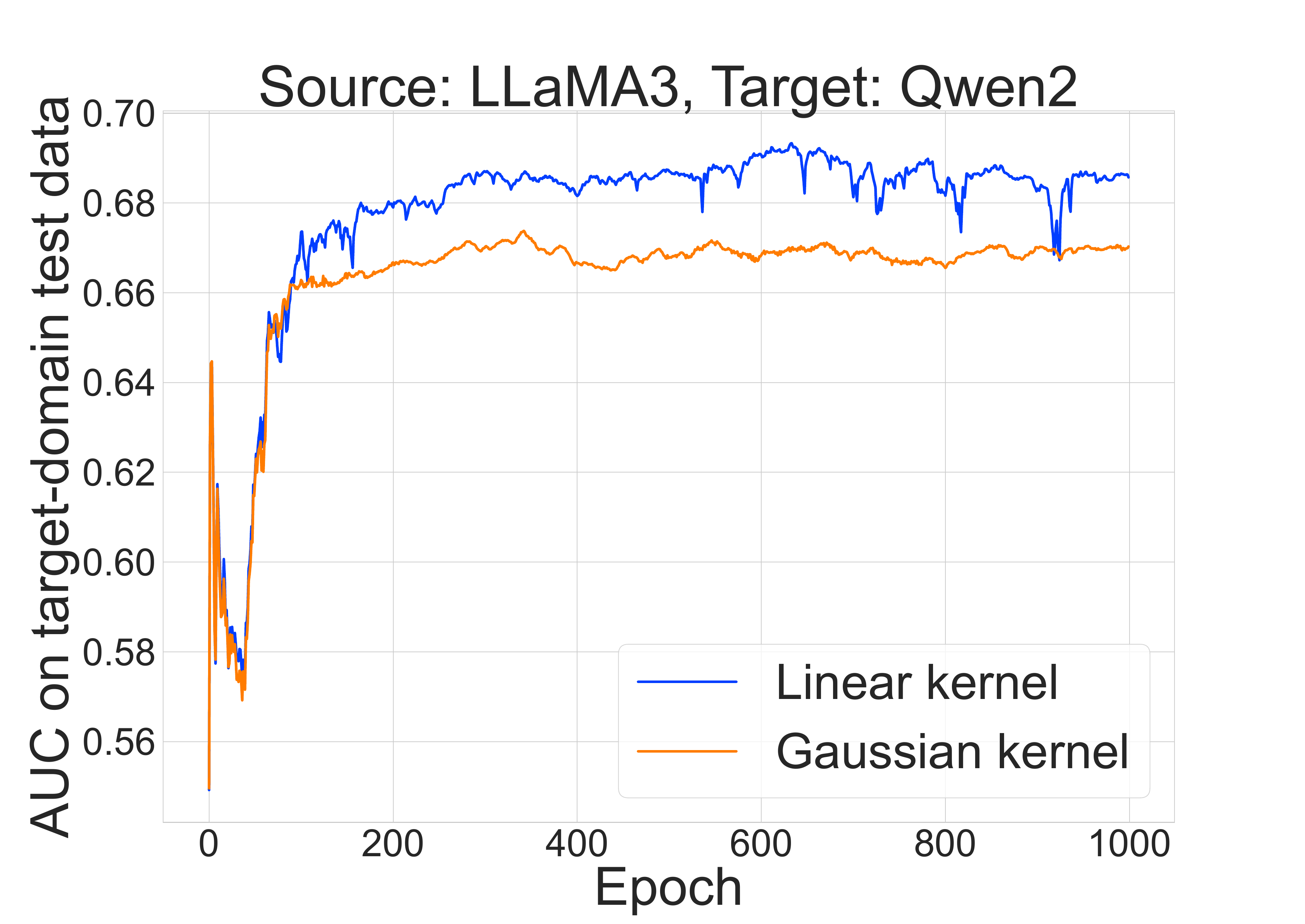}
        \includegraphics[width=1.85in]{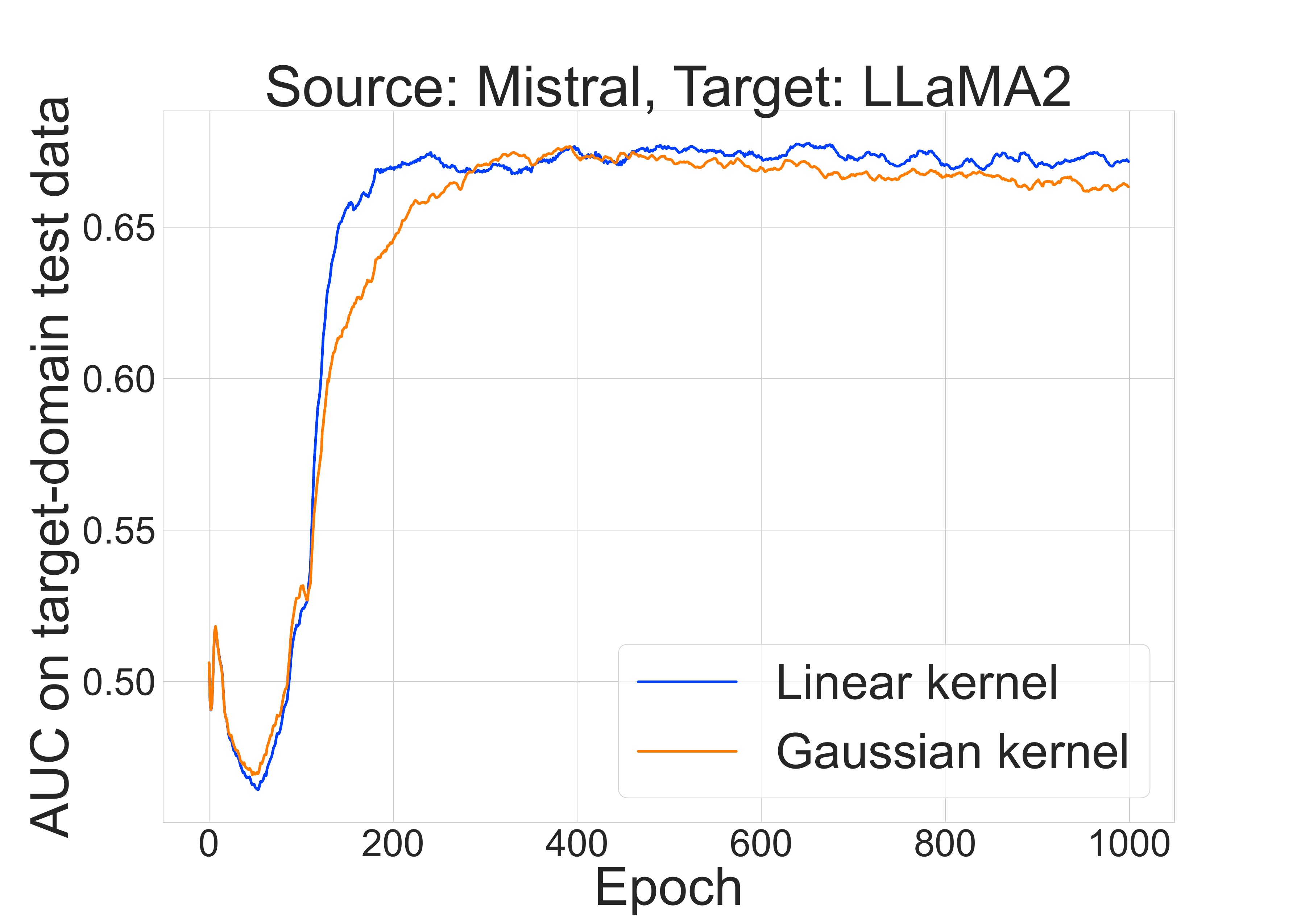}
        \includegraphics[width=1.85in]{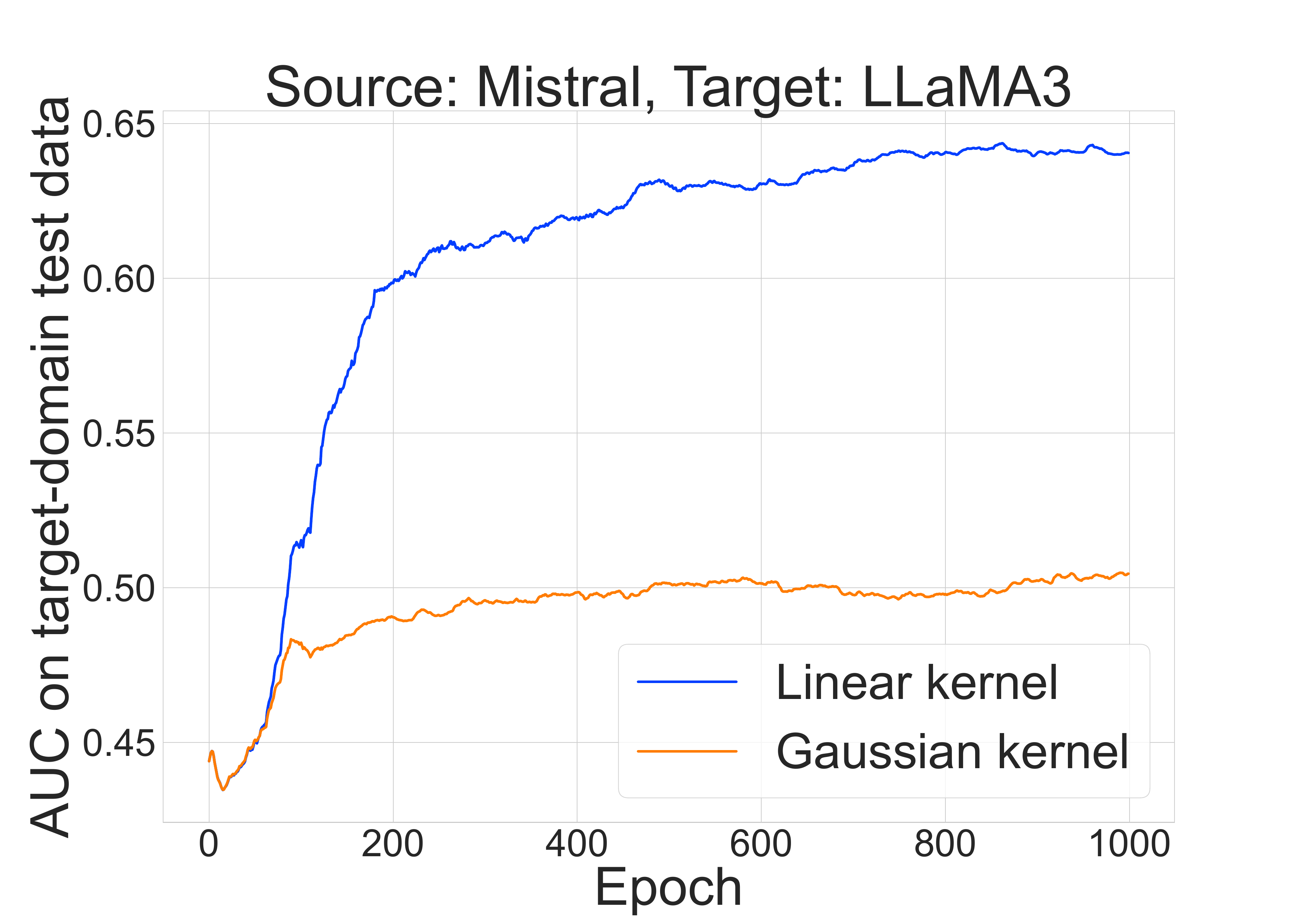}
        \includegraphics[width=1.85in]{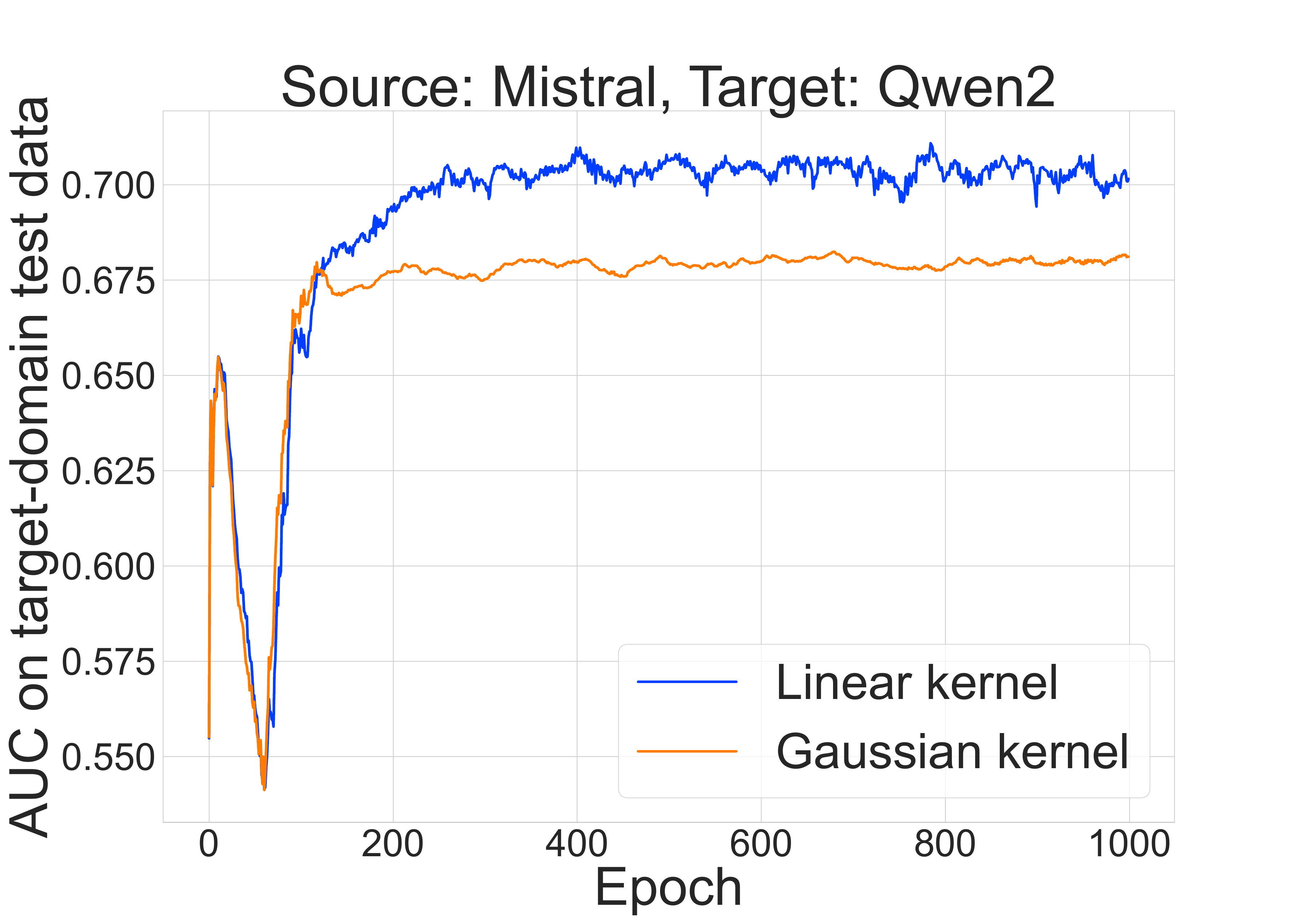}
        \includegraphics[width=1.85in]{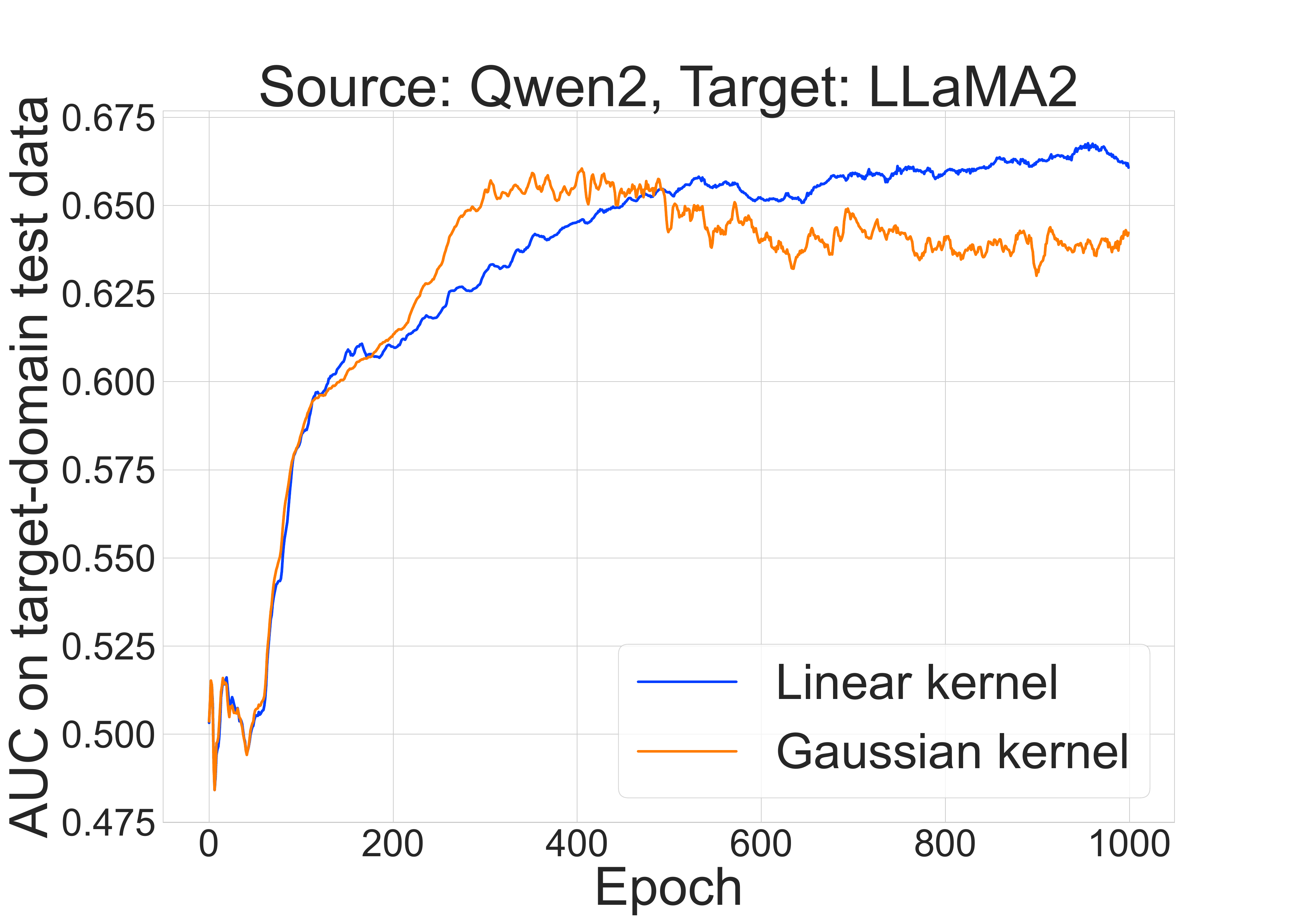}
        \includegraphics[width=1.85in]{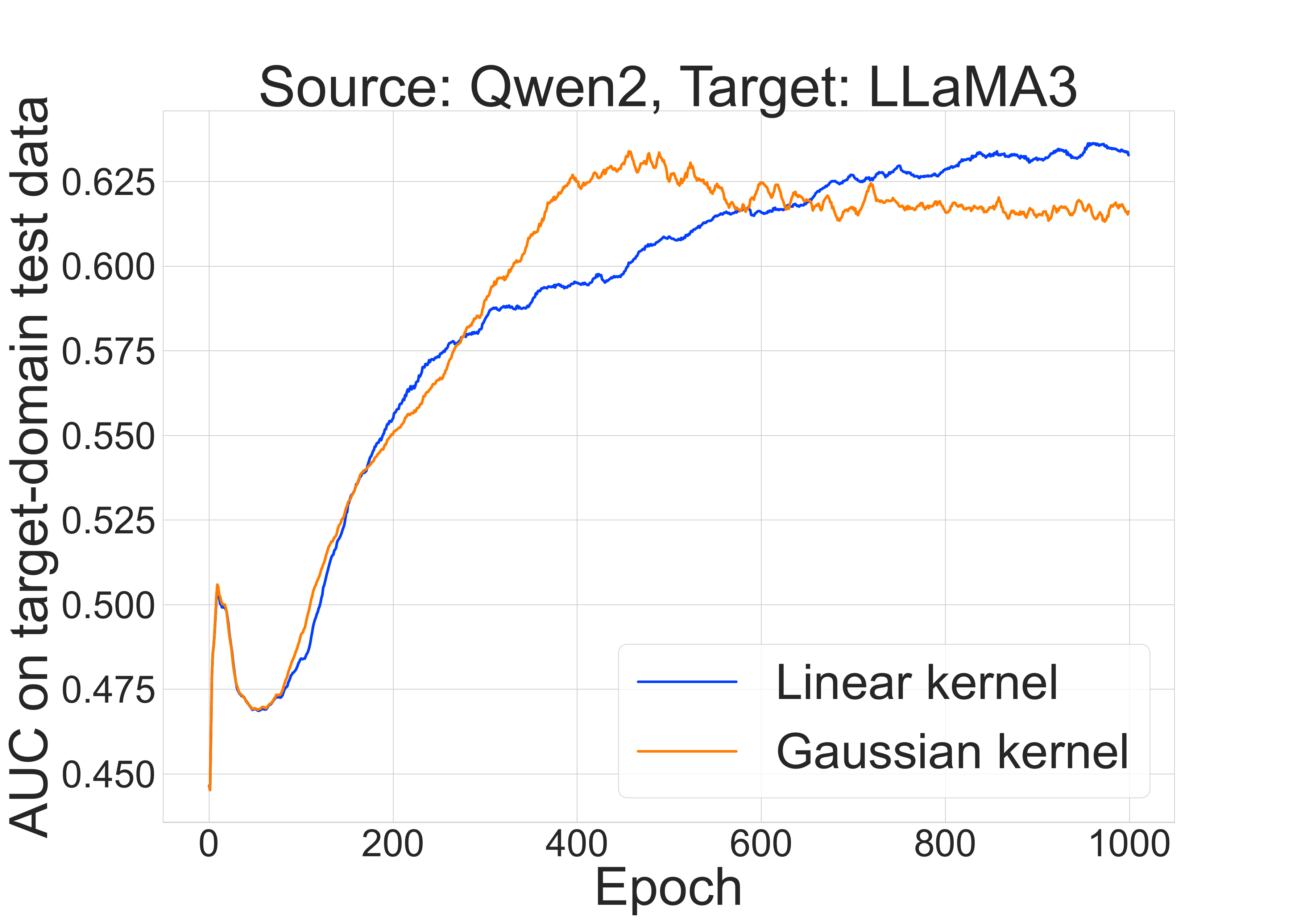}
            \includegraphics[width=1.85in]{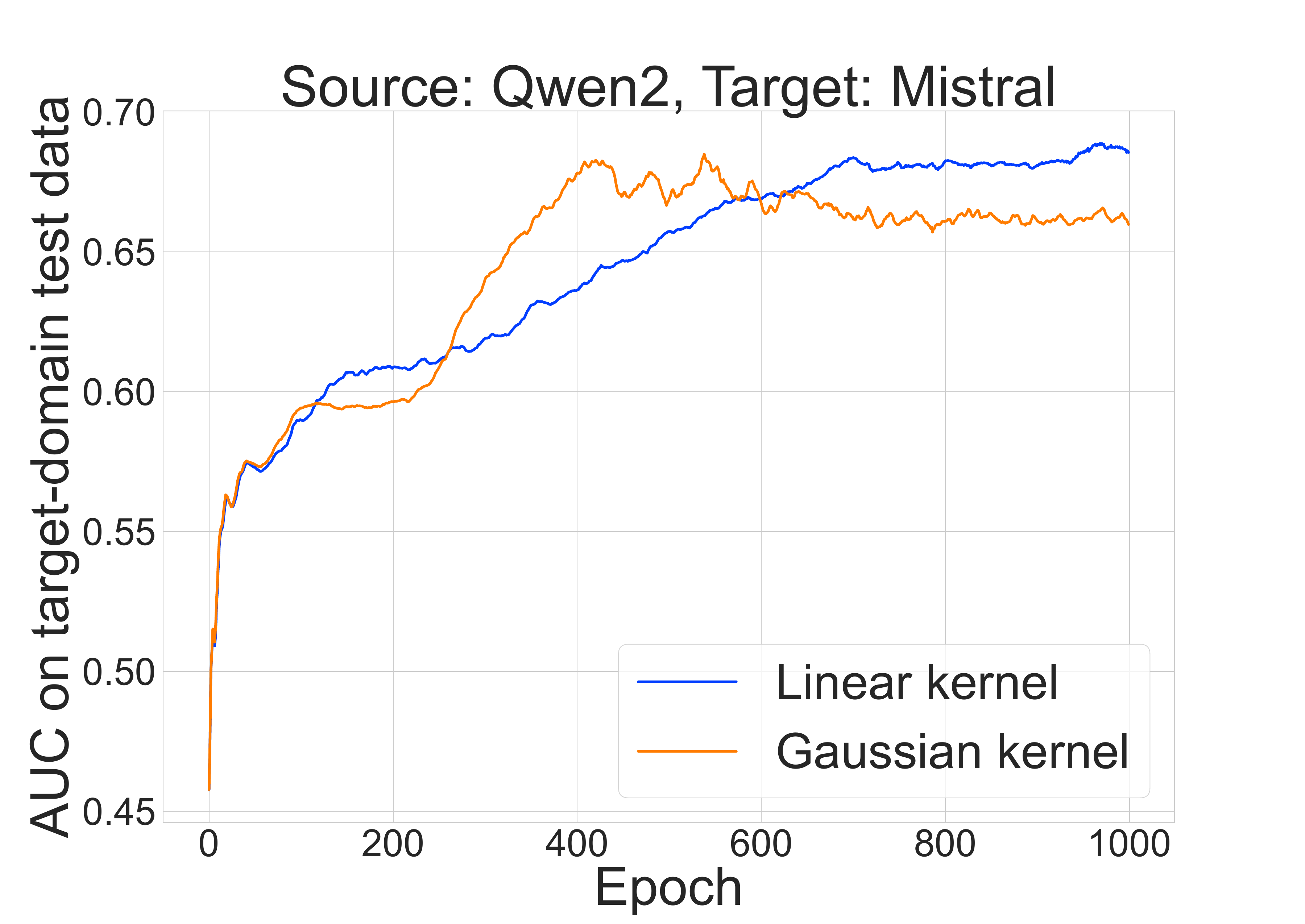}

    \caption{Evaluation of kernels used by MMD loss for training cross-LLM \model. The results are derived from the NQ dataset. Similar trends are observed on other QA datasets.}
    \label{fig:transfer_performance_kernel_selection}
\end{figure*}

\subsection{Kernel Selection for MMD Loss}\label{appendix:kernel selection}
In the MMD loss, the data features are mapped into a reproducing kernel Hilbert space (RKHS) determined by a kernel function. Then distribution distance between different data domains is measured within the RKHS.
We minimize the MMD loss to find a domain-invariant NFP feature space.
Here, we evaluate two commonly employed kernel functions: the linear kernel and the Gaussian kernel~\citep{MMD}. As depicted in Figure~\ref{fig:transfer_performance_kernel_selection}, the linear kernel tends to perform better. This suggests that the features extracted by $g_{enc}$ for NFP tasks are already inherently discriminative.

\section{Demo \& Case Study}\label{appendix:case_study}
\textbf{Human Evaluation via Demo.} We have implemented a demo of \model.
In the demo, a user can choose a specific LLM, and then enter a fact-seeking question in the text box.
After submitting the question, \smodel will return whether the LLM knows the factual answer.
Then the user can decide whether to call the LLM to generate the response. If the user decides to query the LLM, the demo will provide the response generated by the LLM. 
According to the prediction of \smodel and the LLM's response, the user can score the performance of \model.
We recruited 22 volunteers, consisting of 11 females and 11 males with bachelor's degrees or higher, to use our demo and rate its performance.
3 points indicate that the prediction of \smodel is correct, 2 points indicate that \smodel acknowledges its lack of confidence in the prediction result, and 1 point indicates that the prediction of \smodel is incorrect.
We received 680 de-duplicated user queries, with 127 (18.7\%) receiving 1 point, 70 (10.3\%) receiving 2 points, and 483 (71.0\%) receiving 3 points.

\textbf{\smodel in the Demo.} Taking LLaMA2-7B-Chat (abbreviated as LLaMA2) as the example, we integrate its NFP datasets, i.e., LLaMA2-PQ, LLaMA2-EQ, and LLaMA2-NQ, to train a \model. Specifically, we use instances from LLaMA2-PQ, and LLaMA2-EQ for training, and instances from LLaMA2-NQ for validation. 
That is because NQ, released by Google, consists of questions posed by real users.
We set the learning rate to 1e-3, and determine the training epochs based on the performance on the validation set. On the validation set, we use \smodel to predict the probability $\mathbf{p}\left(y=1|\mathbf{x}\right)$ for each instance. Then we calculate the averaged probability $\mathbf{\overline{p}}\left(y=1|\mathbf{x}\right)_{pos}$ based on positive instances in the validation set, as well as the averaged probability $\mathbf{\overline{p}}\left(y=1|\mathbf{x}\right)_{neg}$ based on negative instances in the validation set. $\mathbf{\overline{p}}\left(y=1|\mathbf{x}\right)_{pos}$ and $\mathbf{\overline{p}}\left(y=1|\mathbf{x}\right)_{neg}$ are used as the thresholds for predicting whether the LLM knows the factual answers. Clearly, given a question, \smodel predicts the probability of LLaMA2 generating a non-factual response. If the probability is larger than $\mathbf{\overline{p}}\left(y=1|\mathbf{x}\right)_{pos}$, the demo outputs ``The LLM does not know the factual answer''; If the probability is smaller than $\mathbf{\overline{p}}\left(y=1|\mathbf{x}\right)_{neg}$, the demo outputs ``The LLM knows the factual answer''; otherwise, the demo outputs ``I am not sure if the LLM knows the factual answer''.
Similarly, we train \smodel for LLaMA3-8B-Instruct, Mistral-7B-Instruct-v0.2, and Qwen2-1.5B-Instruct, respectively.

\textbf{Case Study and Future Work.}
Taking LLaMA2-7B-Chat as an example, we show some good cases at the end of this paper, based on questions from our human evaluation. These questions cover a range of topics, such as movies, TV series, music, brands, games, and science.
We observe that \smodel works well on most questions, and it is more likely to mistakenly predict on questions related to periodic events.
The recurrence of similar events could impact \smodel's judgments. 
In other words, LLMs could have limited self-awareness in recognizing the nuances of periodic events.
We plan to enhance \model’s ability to better handle such questions in future work.

\clearpage

\onecolumn 

\begin{tcolorbox}[colframe=gray!50!black,colback=gray!10!white,boxrule=1pt,boxsep=0pt,left=5pt,right=5pt,fontupper=\footnotesize, width=1\textwidth, title={\parbox{\textwidth}{\raggedright \textbf{\textcolor{green}{Good Case}}}}]
\textbf{User Question:}

Who was the screenwriter for The Toy?

\vspace{1.5em}

\textbf{Prediction from \model:}

The LLM {does not know} the factual answer.

\vspace{1.5em}

\textbf{Main Response from LLM:}

The screenwriter for the 1982 film ``The Toy'' is:
\\
\\* Tom Mankiewicz

\vspace{1.5em}

\textbf{Note:}

The LLM's response is incorrect.
The screenwriter is not Tom Mankiewicz.
\end{tcolorbox}

\vspace{1em} 

\begin{tcolorbox}[colframe=gray!50!black,colback=gray!10!white,boxrule=1pt,boxsep=0pt,left=5pt,right=5pt,fontupper=\footnotesize, width=1\textwidth, title={\parbox{\textwidth}{\raggedright \textbf{\textcolor{green}{Good Case}}}}]
\textbf{User Question:}

Who is the director of TV series The Happy Family?

\vspace{1.5em}

\textbf{Prediction from \model:}

The LLM {does not know} the factual answer.

\vspace{1.5em}

\textbf{Main Response from LLM:}

I apologize, but I don't have access to information on the director of a specific TV series
called ``The Happy Family.'' Could you please provide more context or details about the
series, such as its release date, network, or any other relevant information? This will
help me provide a more accurate answer.

\vspace{1.5em}

\textbf{Note:}

The LLM acknowledges that it does not know the factual answer.
\end{tcolorbox}

\vspace{1em} 

\begin{tcolorbox}[colframe=gray!50!black,colback=gray!10!white,boxrule=1pt,boxsep=0pt,left=5pt,right=5pt,fontupper=\footnotesize, width=1\textwidth, title={\parbox{\textwidth}{\raggedright \textbf{\textcolor{green}{Good Case}}}}]
\textbf{User Question:}

Who wrote the lyrics for the song I Can Do It With A Broken Heart?

\vspace{1.5em}

\textbf{Prediction from \model:}

The LLM {does not know} the factual answer.

\vspace{1.5em}

\textbf{Main Response from LLM:}

I apologize, but I don't have access to the lyrics of the song ``I Can Do It With A Broken
Heart'' as it may be a private or unreleased song. Additionally, I cannot provide lyrics for
songs that are not publicly available or have not been released.

\vspace{1.5em}

\textbf{Note:}

The LLM acknowledges that it does not know the factual answer.
\end{tcolorbox}

\vspace{1em} 

\begin{tcolorbox}[colframe=gray!50!black,colback=gray!10!white,boxrule=1pt,boxsep=0pt,left=5pt,right=5pt,fontupper=\footnotesize, width=1\textwidth, title={\parbox{\textwidth}{\raggedright \textbf{\textcolor{green}{Good Case}}}}]
\textbf{User Question:}

Which TV series does the character Sheldon Lee Cooper come from?

\vspace{1.5em}

\textbf{Prediction from \model:}

The LLM {knows} the factual answer.

\vspace{1.5em}

\textbf{Main Response from LLM:}

The character Sheldon Lee Cooper is from the popular TV series ``The Big Bang Theory.''

\vspace{1.5em}

\textbf{Note:}

The LLM's response is correct.
\end{tcolorbox}

\vspace{1em} 

\begin{tcolorbox}[colframe=gray!50!black,colback=gray!10!white,boxrule=1pt,boxsep=0pt,left=5pt,right=5pt,fontupper=\footnotesize, width=1\textwidth, title={\parbox{\textwidth}{\raggedright \textbf{\textcolor{green}{Good Case}}}}]
\textbf{User Question:}

Who founded KFC?

\vspace{1.5em}

\textbf{Prediction from \model:}

The LLM {knows} the factual answer.

\vspace{1.5em}

\textbf{Main Response from LLM:}

KFC, or Kentucky Fried Chicken, was founded by Colonel Harland Sanders.

\vspace{1.5em}

\textbf{Note:}

The LLM's response is correct.
\end{tcolorbox}

\vspace{1em} 

\begin{tcolorbox}[colframe=gray!50!black,colback=gray!10!white,boxrule=1pt,boxsep=0pt,left=5pt,right=5pt,fontupper=\footnotesize, width=1\textwidth, title={\parbox{\textwidth}{\raggedright \textbf{\textcolor{green}{Good Case}}}}]
\textbf{User Question:}

Which is heavier, 1kg iron or 1kg cotton?

\vspace{1.5em}

\textbf{Prediction from \model:}

The LLM {knows} the factual answer.

\vspace{1.5em}

\textbf{Main Response from LLM:}

Both 1 kg of iron and 1 kg of cotton weigh the same amount, 1 kilogram.

\vspace{1.5em}

\textbf{Note:}

The LLM's response is correct.
\end{tcolorbox}

\vspace{1em} 

\begin{tcolorbox}[colframe=gray!50!black,colback=gray!10!white,boxrule=1pt,boxsep=0pt,left=5pt,right=5pt,fontupper=\footnotesize, width=1\textwidth, title={\parbox{\textwidth}{\raggedright \textbf{\textcolor{orange}{Bad Case}}}}]
\textbf{User Question:}

Which team won the women’s volleyball championship at the 2004 Athens Olympics?

\vspace{1.5em}

\textbf{Prediction from \model:}

The LLM {knows} the factual answer.

\vspace{1.5em}

\textbf{Main Response from LLM:}

The United States women's volleyball team won the gold medal at the 2004 Athens Olympics. 

\vspace{1.5em}

\textbf{Note:}

The LLM's response is incorrect. The correct answer is the Chinese women's volleyball team.
\end{tcolorbox}

\vspace{1em} 

\begin{tcolorbox}[colframe=gray!50!black,colback=gray!10!white,boxrule=1pt,boxsep=0pt,left=5pt,right=5pt,fontupper=\footnotesize, width=1\textwidth, title={\parbox{\textwidth}{\raggedright \textbf{\textcolor{orange}{Bad Case}}}}]
\textbf{User Question:}

Who won the Turing Award in 2016?

\vspace{1.5em}

\textbf{Prediction from \model:}

The LLM {knows} the factual answer.

\vspace{1.5em}

\textbf{Main Response from LLM:}

The Turing Award for 2016 was awarded to Yoshua Bengio, Geoffrey Hinton, and Yann LeCun for their contributions to the development of deep learning.

\vspace{1.5em}

\textbf{Note:}

The LLM's response is incorrect. The correct answer is Tim Berners-Lee.
\end{tcolorbox}

\vspace{1em} 

\begin{tcolorbox}[colframe=gray!50!black,colback=gray!10!white,boxrule=1pt,boxsep=0pt,left=5pt,right=5pt,fontupper=\footnotesize, width=1\textwidth, title={\parbox{\textwidth}{\raggedright \textbf{\textcolor{orange}{Bad Case}}}}]
\textbf{User Question:}

Where was the 2006 World Expo held?

\vspace{1.5em}

\textbf{Prediction from \model:}

The LLM {knows} the factual answer.

\vspace{1.5em}

\textbf{Main Response from LLM:}

The 2006 World Expo was held in Aichi, Japan, from March 25 to September 25, 2005.

\vspace{1.5em}

\textbf{Note:}

The LLM's response is incorrect. The correct answer is Chiang Mai, Thailand.
\end{tcolorbox}

\vspace{1em} 

\begin{tcolorbox}[colframe=gray!50!black,colback=gray!10!white,boxrule=1pt,boxsep=0pt,left=5pt,right=5pt,fontupper=\footnotesize, width=1\textwidth, title={\parbox{\textwidth}{\raggedright \textbf{\textcolor{orange}{Bad Case}}}}]
\textbf{User Question:}

Who won the Nobel Prize in Physics in 1903?

\vspace{1.5em}

\textbf{Prediction from \model:}

The LLM {knows} the factual answer.

\vspace{1.5em}

\textbf{Main Response from LLM:}

The Nobel Prize in Physics was not awarded in 1903. The Nobel Prize in Physics was first awarded in 1901, and the first Nobel Prize ceremony was held in 1904. Therefore, there was no Nobel Prize in Physics awarded in 1903.

\vspace{1.5em}

\textbf{Note:}

The LLM's response is incorrect. The correct answer is Henri Becquerel, Pierre Curie, and Marie Curie.
\end{tcolorbox}

\label{sec:appendix}

\end{document}